\documentclass{article}

\usepackage{microtype}
\usepackage{graphicx}
\usepackage{subfigure}
\usepackage{booktabs} 

\usepackage{amsmath}
\usepackage{etoolbox}

\usepackage{amssymb}
\usepackage{algorithmic}

\usepackage{forloop}
\usepackage{fancyhdr}
\usepackage{stmaryrd}
\usepackage{textcomp}
\usepackage{bm}

\usepackage{amssymb}
\usepackage{booktabs}
\usepackage{kotex}
\usepackage{tablefootnote}
\usepackage{footnote}
\usepackage[dvipsnames,table,xcdraw]{xcolor}
\usepackage{hyperref}
\usepackage{graphicx}
\usepackage{ulem} 
\usepackage{textcomp}
\usepackage{latexsym}
\usepackage{graphicx}
\usepackage{tabularx}
\usepackage{graphics}
\usepackage{adjustbox}

\usepackage[dvipsnames]{xcolor}
\usepackage[linesnumbered,ruled,algo2e]{algorithm2e}
\usepackage{lineno,hyperref}
\usepackage{amssymb}
\usepackage{amsmath}
\usepackage{array}
\usepackage{hhline}
\usepackage{setspace}
\usepackage{cleveref}
\usepackage{ulem}
\usepackage{amssymb}
\usepackage{pifont}
\usepackage{cancel}
\usepackage{bbm}
\usepackage{multirow}
\usepackage{subfigure}

\usepackage{graphicx}
\usepackage{tabularx}
\usepackage{graphics}
\usepackage{adjustbox}

\usepackage[english]{babel}
\usepackage[utf8]{inputenc}

\usepackage{relsize}
\usepackage{mathtools}
\usepackage{standalone}
\usepackage{tcolorbox}
\usepackage{CJK}
\usepackage{adjustbox}

\usepackage{xcolor}

\SetCommentSty{mycommfont}

\newcolumntype{C}[1]{>{\centering\arraybackslash\hspace{0pt}}m{#1}}
\newcolumntype{R}[1]{>{\raggedleft\arraybackslash}m{#1}}

\newcommand{\RNum}[1]{\lowercase\expandafter{\romannumeral #1\relax}}

\modulolinenumbers[5]

\graphicspath{{./figure/}}

\usepackage[accepted]{icml2021}

\icmltitlerunning{LAnoBERT: System Log Anomaly Detection based on BERT Masked Language Model}

\begin{document}








\twocolumn[
\icmltitle{LAnoBERT : System Log Anomaly Detection \\based on BERT Masked Language Model}

\icmlsetsymbol{equal}{*}

\begin{icmlauthorlist}
\icmlauthor{Yukyung Lee}{Korea}
\icmlauthor{Jina Kim}{Korea}
\icmlauthor{Pilsung Kang}{Korea}
\end{icmlauthorlist}

\icmlaffiliation{Korea}{School of Industrial Management Engineering,
College of Engineering,
Korea University, Seoul, Korea}

\icmlcorrespondingauthor{Pilsung Kang}{pilsung\_kang@korea.ac.kr}

\icmlkeywords{Log data analysis, anomaly detection, Transformer, BERT}

\vskip 0.3in]



\printAffiliationsAndNotice{ } 

\begin{abstract}
The system log generated in a computer system refers to large-scale data that are collected simultaneously and used as the basic data for determining errors, intrusion and abnormal behaviors. The aim of system log anomaly detection is to promptly identify anomalies while minimizing human intervention, which is a critical problem in the industry. Previous studies performed anomaly detection through algorithms after converting various forms of log data into a standardized template using a parser. Particularly, a template corresponding to a specific event should be defined in advance for all the log data using which the information within the log key may get lost. In this study, we propose LAnoBERT, a parser free system log anomaly detection method that uses the BERT model, exhibiting excellent natural language processing performance. The proposed method, LAnoBERT, learns the model through masked language modeling, which is a BERT-based pre-training method, and proceeds with unsupervised learning-based anomaly detection using the masked language modeling loss function per log key during the test process. In addition, we also propose an efficient inference process to establish a practically applicable pipeline to the actual system. Experiments on three well-known log datasets, i.e., HDFS, BGL, and Thunderbird, show that not only did LAnoBERT yield a higher anomaly detection performance compared to unsupervised learning-based benchmark models, but also it resulted in a comparable performance with supervised learning-based benchmark models.
\end{abstract}

\section{Introduction}
\label{sec:Introduction}

Owing to the recent advancement of the IT industry, a growing emphasis is placed on the importance of the system log data for identifying problems when accidents or failures occur in programs \cite{He2017DrainAO}. The system log comprises large-scale data collected simultaneously in a computer system and used as the basic data for determining anomalies; thus it is a very critical and valuable resource. The log data generated from various systems should be monitored in real-time for system stability because they represent the current status of a system. Real-time monitoring is conventionally performed by operators; however, such a method entails the possibility of including errors and bias depending on the operator and is limited by being unable to promptly detect system anomalies \cite{1607498}. Subsequently, anomaly detection using rule-based algorithms has been proposed to reduce human error \cite{6320555}. However, rule-based methodologies also require human intervention; therefore, research is being actively conducted on real-time monitoring-based anomaly detection methods based on machine learning, which minimizes human intervention \cite{10.1145/3133956.3134015}.

Log data are sequence data collected in real-time. They consist of a combination of log keys, which can be considered as words, whereas log sequences can be considered as sentences; a log sequence is generated through a series of syntax rules \cite{7837916}. Also, since log data is accumulated based on user actions at regular time intervals, there are many duplicates in an actual log history. Hence, although the total amount of log instances is very large, a single log sequence is short and the number of unique log keys are limited in general.

Machine learning-based log anomaly detection involves three steps: 1) preprocessing log keys, 2) feature embedding, and 3) anomaly detection. \textbf{Preprocessing of log keys} refers to refining unstructured log keys and can be performed with or without a log parser. Parsing-based log anomaly detection involves generating log data in a standardized template format using a log parser. \textbf{Feature embedding} involves extracting features from preprocessed log sequences. Recent methods \cite{nedelkoski2020self} use transformer-based models, whereas earlier methods used RNNs to treat log sequences as natural language \cite{10.1145/3217871.3217872, 10.1145/3133956.3134015, DBLP:journals/corr/KimYLPY16}. \textbf{Anomaly detection} involves finding abnormal logs using the extracted features.

Previous log anomaly detection studies \cite{10.1145/3133956.3134015, zhang2019robust, huang2020hitanomaly, nedelkoski2020self} showed remarkable performance on open datasets, but have limitations in terms of practicality and extensibility. 
\begin{itemize}
    \item \textbf{Reliance on Log Parsers}: Parser-based log preprocessing requires predefined templates for standardizing log keys and manual refinement by experts \cite{7837916}. This method may result in loss of crucial information during standardization \cite{huang2020hitanomaly} and its performance becomes dependent on log parser compatibility rather than the logic of anomaly detection models \cite{nedelkoski2020self}.
    \item \textbf{Feature embedding with rich semantics for long-term dependency}: In the field of log anomaly detection, previous feature embedding methods primarily utilized RNN-based algorithms, which have been successful in natural language processing \cite{NIPS2017_3f5ee243}. However, these algorithms have difficulties in handling long sequences, particularly with regard to modeling long-term dependencies. To address these limitations, current research is exploring the use of transformer-based architecture \cite{nedelkoski2020self}, which has recently demonstrated exceptional performance in natural language processing \cite{NIPS2017_3f5ee243}.
    \item \textbf{Unrealistic problem formulation}: In prior research, log anomaly detection was mostly formulated as a binary classification instead of anomaly detection \cite{huang2020hitanomaly, nedelkoski2020self, zhang2019robust}. However, in practical systems, the majority of logs are normal, with only a small amount of abnormal logs. Formulating log anomaly detection as a binary classification problem requires a sufficient amount of abnormal data for model training, which is unrealistic as abnormal data is rare in real-world systems. Hence, a more practical approach is to train the model using only normal log data and then utilize abnormal data only during testing, better reflecting real-world scenarios.
\end{itemize}

As a solution for the aforementioned problem, this study proposes a new log anomaly detection model (\textbf{LAnoBERT}; \underline{\textbf{L}}og \underline{\textbf{An}}omaly detection based on \underline{\textbf{BERT}}) established on the following three improvement plans. In LAnoBERT, a simple preprocessing approach utilizing regular expressions was selected to mitigate information loss during the parsing process and to minimize dependence on a specific log parser. Contextualized embedding was extracted using the BERT model, which was trained from scratch to learn the log key sequences, in contrast to previous models which relied on static embedding for feature extraction. Lastly, unsupervised learning-based anomaly detection was performed under the assumption that the context of normal logs differs from that of abnormal logs. In the proposed model, LAnoBERT, masked language modeling of BERT \cite{devlin-etal-2019-bert} was utilized to perform anomaly detection based on the masking predictive probability. An efficient inference process was also proposed, where a log dictionary database was defined, and log key matching was performed for anomaly detection. The model demonstrated superior performance compared to previous models on benchmark datasets of system logs (HDFS, BGL, and Thunderbird). It showed the best performance among unsupervised learning-based models and comparable performance to supervised learning-based models, despite being trained in a less advantageous environment. LAnoBERT satisfied both detection performance and practicality by outperforming some supervised learning-based models.

In summary, the main contributions of our study are as follows.
\begin{itemize}
\item We propose LAnoBERT, a new BERT-based unsupervised and log parser-free anomaly detection framework for log data. Unlike previous studies, it is a log parser-free and unsupervised learning-based model.
\item To improve efficiency, an inference process utilizing a log dictionary database is proposed to identify abnormal logs. This reduces the computational burden of BERT and handles log sequences with lots of redundant information.
\item Despite being trained under less favorable conditions, LAnoBERT demonstrated better or comparable performance to supervised learning-based models. In addition, LAnoBERT effectively detects anomalies in various types of logs, validating its practical usefulness.
\end{itemize}

This paper is organized as follows. In Section \ref{sec:Related Work}, previous studies are reviewed by categorizing them based on neural networks with parsing and free of parsing. In Section \ref{sec:Background}, the background knowledge related to the research is introduced in addition to the log parser and BERT model. Section \ref{sec:Proposed Method} explains the proposed model, LAnoBERT, and its structure, whereas Section \ref{sec:Experimental Setting} describes the experimental design, and Section \ref{sec:Results} describes the log anomaly detection performance. Lastly, in Section \ref{sec:Conclusion}, the conclusion of this study and future research subjects are explained.

\section{Related Work}
\label{sec:Related Work}

Log anomaly detection refers to a method for detecting abnormal logs from a large log dataset. Studies in earlier years \cite{6320555, hansen1993automated, oprea2015detection,prewett2003analyzing, yen2013beehive} performed anomaly detection by regarding specific parts of log data as abnormal. However, these studies had a critical limitation of requiring professional domain-specific knowledge. Recent methodologies involve extracting log data features using a neural network-based model and performing anomaly detection. Log data are unstructured data with a highly complex structure; thus, log anomaly detection can be divided into parsing-based or parsing-free depending on the preprocessing method for the log data. 

\subsection{Parsing-based log anomaly detection}
\label{sec:Parsing-based log anomaly detection}

In these methods, a log parser is needed to perform log anomaly detection. The \textit{Drain}  parser\cite{He2017DrainAO}, which is the most commonly used parser, classifies log messages based on the length using a decision tree and then allocates a log template by exploring the word similarity. In this process, the similarity between the new log message and the existing log template is calculated, and a new template is allocated for the data with a different format from the existing template.

DeepLog \cite{10.1145/3133956.3134015} is the first neural network-based log anomaly detection model in which an anomaly is detected using an unsupervised learning-based LSTM model. In the training phase, a log is generated in a standardized template using the \textit{Drain} parser and then a normal log template pattern is learned. In the test phase, logs having a pattern not trained with the normal data are determined as anomalies. In other words, log patterns with low frequencies are given low scores. The concept of the ‘top g candidate’ is introduced to discern log patterns where it is regarded as normal if the log patterns are present within the candidate or abnormal if not present; thus, the performance varies depending on the candidate.

LogRobust \cite{zhang2019robust} is an attention-based bi-LSTM model for detecting anomalies. After creating a log with a standardized template using the \textit{Drain} parser, the log data features are extracted by generating TF-IDF and a word semantic vector. Because this particular model detects anomalies based on classification, it can be considered as a classification problem. 

HitAnomaly \cite{huang2020hitanomaly} is an anomaly detection model using a transformer. It also conducts preprocessing through the \textit{Drain} parser. A template is standardized through a log parser and the information substituted with a template is defined as parameters. The substituted information refers to the data that get lost without inclusion in the template. Two types of information are separately encoded using a transformer encoder, and two types of representation are combined based on attention to detect anomalies through classification. This model also performs classification-based anomaly detection, thus entailing limitations. 

LogBERT \cite{9534113} is a BERT-based framework for log data anomaly detection, utilizing a \textit{Drain} parser for log sequence refinement. It follows a similar approach to DeepLog in detecting outliers but instead trains using only normal log data through two tasks. The first task, masked log key prediction (MLKP), trains normal log patterns via the same objective function as masked language modeling. The second task, Volume of Hypersphere Minimization (VHM), aims to find the smallest sphere that contains normal logs. In the inference stage, the top $g$ predicted log keys are selected as a candidate set from a randomly masked normal log sequence, and the observed log key is considered as an anomaly if it does not belong to the candidate set. The model detects anomalies by applying BERT's masked language modeling, however, it has a limitation in that it cannot fully consider the log sequence when masking due to the random selection of a log key from the sequence.

In this study, we propose a log anomaly detection model that does not depend on the log parser. Therefore, even when a new log sequence is recorded, data is not parsed using the log template, but the log sequence is refined using simple preprocessing logic. This method can preserve the log sequence as much as possible by minimizing information loss commonly occurred in the parsing process.

\subsection{Parsing-free log anomaly detection}
\label{sec:Parsing-free log anomaly detection}

LogSy \cite{nedelkoski2020self} is a transformer-based anomaly detection model that uses a tokenizer to preprocess log values; thus, it is free from the use of a log parser when detecting anomalies. In LogSy, classification is performed using normal data of a training log and abnormal log data generated from a different system. In addition, training is performed so that the distance between the normal and abnormal log increases through a distance-based loss function. It is different from LogRobust and HitAnomaly because it does not learn the normal and abnormal log generated in one system based on classification like the previous models. Hence, this model also entails various limitations to be used in the industry as it adapts a classification-based approach. 

Additionally, NeuralLog \cite{le2021log} is also a parser-free and classification-based anomaly detection model. While NeuralLog shares a similar structure with LogSy, it distinguishes itself by employing both normal and abnormal data from the target system, as well as a separate system, during the training process. In contrast, LogSy addresses the classification problem by relying solely on normal data from the target system and abnormal data generated from a different system.

The proposed model is also a log parser-free methodology. After refining the log through a simple preprocessing logic, the log sequence is segmented using a word-piece tokenizer. Through this, the log sequence is not categorized into one of the predefined templates, but the log sequence itself is used as an input of the anomaly detection model. Also, it can be flexibly applied even if a new log sequence that has not been processed ever.

\section{Background}
\label{sec:Background}

\subsection{Log parser for anomaly detection}
\label{sec:Log parser for anomaly detection}

Log data are large-scale data collected in real-time, and raw log messages are usually unstructured because developers are allowed to write free-text log messages in the source code \cite{He2017DrainAO}. Therefore, a log sequence is unstructured data that must be converted to structured data. A log parser \cite{7837916, He2017DrainAO} is a technique proposed for this process. When a standardized log template is generated from an actual log, highly complicated data are simply preprocessed for substitution with very few events. For example, 4,747,964 log messages generated from the BGL system are converted to 376 events through the \textit{Drain} parser \cite{He2017DrainAO}. Then, when anomaly detection is performed with preprocessed data, the anomalies can be detected through a simple process. However, the performance of log anomaly detection models using a parser becomes heavily dependent on a log parser \cite{nedelkoski2020self}. 

\subsection{BERT}
\label{sec:BERT}

BERT \cite{devlin-etal-2019-bert} is a model consisting of a transformer \cite{NIPS2017_3f5ee243} encoder, which achieved outstanding performance in various natural language processing tasks \cite{devlin-etal-2019-bert}. One of the major characteristics of BERT is that pre-training is performed using two unsupervised learning methods, which are masked language modeling (MLM) and next sentence prediction (NSP). MLM involves replacing certain tokens of an input sentence with ‘[MASK]’ and predicting that they would appear in the corresponding position. NSP involves combining two sentences with the token ‘[SEP]’ in between, and then predicting whether the two sentences are semantically connected through the ‘[CLS]’ token positioned in the very front of the input sentence. These two tasks do not require labeled data as in a specific downstream task; thus, general-purpose knowledge can be sufficiently learned through pre-training using a massive unlabeled dataset \cite{clark-etal-2019-bert, jawahar-etal-2019-bert, tenney-etal-2019-bert}. BERT that has been pre-trained is being applied in fields using sequence data in addition to natural language processing; some of the examples include ProtTrans \cite{9477085} and ESM \cite{rao2020transformer}.

\subsection{BERT for anomaly detection}
\label{sec:BERT for anomaly detection}

The system log can be deemed as sequence data because it is a dataset with an order. Therefore, previous methodologies applied the techniques used for natural language processing to extract the features of logs. The system log data encompass both log messages and natural language. In this study, we propose a BERT-based system log anomaly detection system to overcome the limitations of existing methodologies. Previous methodologies treated all log data as sequence data, but applying BERT enables the learning of both the log features and natural language. Moreover, a tokenizer of BERT can be applied without using a separate log parser during which natural language data that are lost while converting to a template using a log parser can be preserved. Additionally, a model capable of capturing the semantics and context of the system log is necessary for accurately detecting abnormal logs in the system log. It is crucial to capture the semantics and context of the system log because the words appearing in the system log may have a different meaning from natural language. The goal of this research is to implement an effective pre-training approach for the system log utilizing masked language modeling in a bi-directional context. Additionally, we present a novel framework for identifying context anomalies by means of the trained models' MLM loss and predictive probability, along with a log key matching technique during the inference stage.

\section{Proposed Method}
\label{sec:Proposed Method}

\begin{figure*}[h!]
\centering
\includegraphics[width=0.8\textwidth]{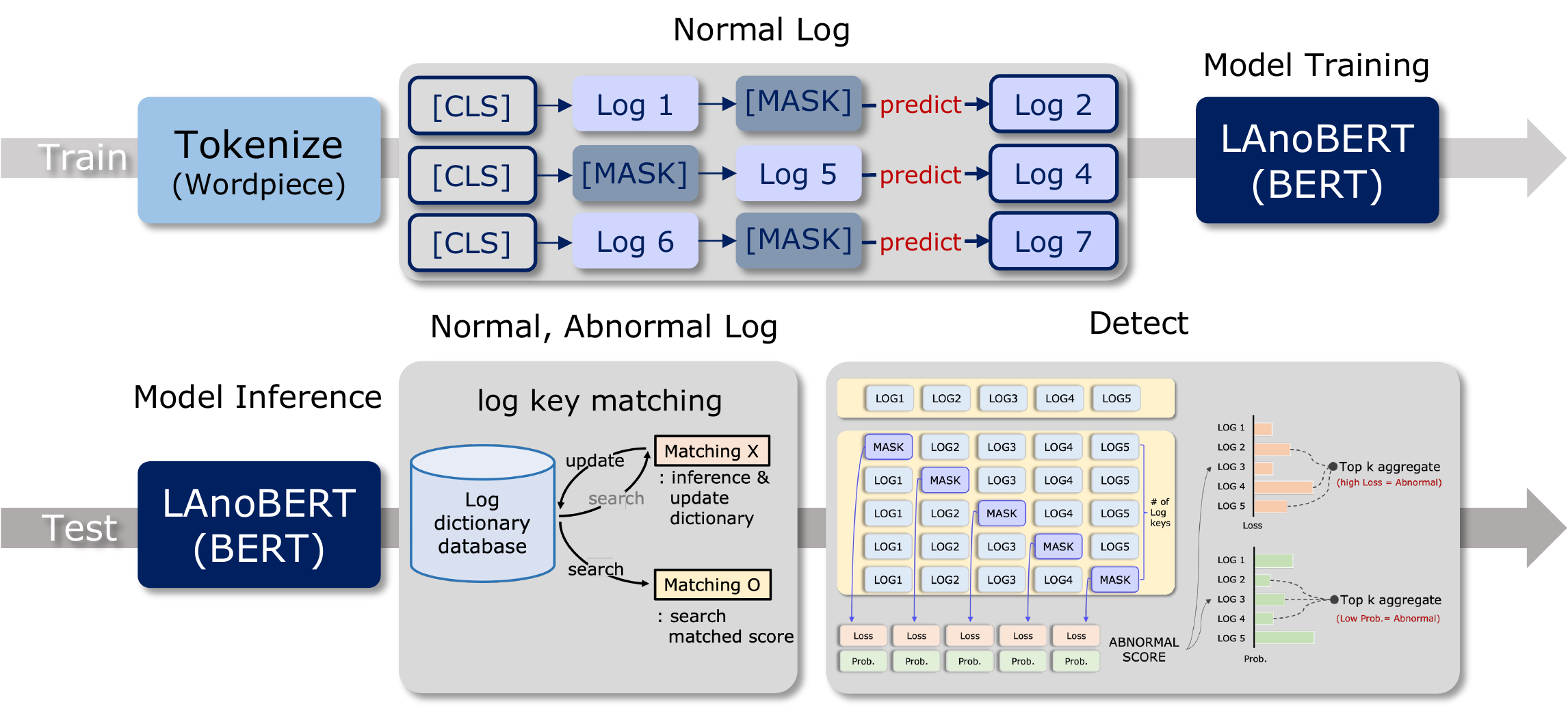} 
\caption{The architecture of LAnoBERT.}
\label{fig:LAnoBERT_Architecture}
\end{figure*}

In this chapter, the major network used in the proposed methodology and the architecture of the proposed model are explained. The description and significance of the MLM of BERT are presented in Section \ref{sec:Masked Language Model}, and the training purpose and execution procedure of the proposed model are presented in Section \ref{sec:Problem Definition} and \ref{sec:LAnoBERT}, respectively.

\subsection{Masked Language Model}
\label{sec:Masked Language Model}

The operation mechanism of LAnoBERT proposed in this study is shown in Figure \ref{fig:LAnoBERT_Architecture}. Because LAnoBERT is executed through MLM, which is a pre-training method of BERT, MLM is explained in detail in this section, and the log anomaly detection procedure is explained in depth in Section \ref{sec:Problem Definition}.

MLM was inspired by the cloze task \cite{taylor1953cloze} where certain tokens of an input sentence are replaced with [MASK] and then the words in the [MASK] tokens are predicted. Entire sentences are replaced with the [MASK] token at an arbitrary probability of 15\%, and appropriate words can be predicted only based on the context. Particularly, the MLM objective function can generate bidirectional representations, unlike the pre-training of left-to-right language models. Therefore, the proposed method can pre-train the deep bidirectional transformer \cite{devlin-etal-2019-bert}.

According to XLNet \cite{yang2019xlnet}, MLM can be defined as an auto-encoding (AE) pre-training object. When $n$ is the sequence length and the $i$-th token is $x_i$, the given input sequence can be expressed as $X=[x_1,x_2,...,x_n]$. If the [MASK] token is defined as $\bar{x}=[MASK]$ , the input sequence containing noise can be expressed as $\hat{X}=[x_1,[MASK],...,x_n]$. In BERT, specific tokens are substituted with the special token [MASK] at a pre-determined probability (15\%). Here, the likelihood and objective function can be expressed as follows.

\begin{equation} 
\label{eq:1}
\begin{split}
   p(\bar{X}|\hat{X})\ \approx\prod_{n=1}^{N}{p(x_n|\hat{X})}
\end{split},
\end{equation}

\begin{equation} \label{eq:2}
\begin{split}
    & {Max}_\theta\ log\ p(\bar{x}|\hat{X})\\
    & \approx \sum_{n=1}^{N}{m_n\ log\ p_\theta(x_n|\hat{X})}\\
    & = {\mathlarger{\sum}_{n=1}^{N}
    {m_n\ log\ \frac{exp(H_\theta({(\hat{x})}_n^\intercal e(x_n))}{\sum_{x\prime}
    {exp(H_\theta({(\hat{x})}_n^\intercal e(x\prime))}}p_\theta(x_n|\hat{X})}}
\end{split}
\end{equation}

In Eq. (2), $m_n$ indicates masking, where $x_n$ is the [MASK] token when $m_n=1$. Furthermore, $H_\theta$ indicates the hidden vector of a transformer encoder. 

In this study, MLM was not only applied in the training phase but also in anomaly detection to detect abnormal logs. The reasons for using MLM in system log anomaly detection are as follows. First, there is ample data available for training BERT because the log data are collected in real-time. Most of the collected data are normal log data, which facilitates the effective pre-training of BERT. When a sufficient number of data is given, BERT can obtain numerous contextual and structural features during pre-training. Therefore, performing anomaly detection using the proposed model is expected to improve the generalization performance of effectively detecting abnormal logs by adequately learning the features of a normal log system. Second, MLM does not require the labeling of tasks and accords with the purpose of anomaly detection where only normal data are used for training. Because anomaly detection is an unsupervised learning-based methodology where only normal data are used for training, it is appropriate for application to cases where normal data are predominantly greater than abnormal data. Since anomaly detection is an unsupervised learning-based approach that does not use label information during the model training, it is more appropriate than a supervised binary classification-based approach where there is an overwhelming amount of normal data. Third, MLM is an appropriate methodology to apply to anomaly detection from the perspective of prompt-based learning \cite{JMLR:v21:20-074, petroni-etal-2019-language, liu2021pretrain, radford2019language, schick-schutze-2021-exploiting}. In contrast to conventional methods that require layers conforming to tasks to perform downstream tasks, it is suitable for finding patterns of log data in anomaly detection by comparing the actual log keys and the generated log keys. Fourth, the context of abnormal log data can be identified if MLM is performed using only normal log data. Normal log data have a very similar form as abnormal log data, but the probability of certain words appearing varies if the context of surrounding words is considered. It was assumed that the MLM predictive probability of abnormal log data is low when anomaly detection is performed using the BERT model trained only with normal log data, and the performance result was relevant.

\subsection{Problem Definition}
\label{sec:Problem Definition}

The system log anomaly detection in this study can be defined as follows. When $s_{len}$ is the sequence length, the log key of the $i$-th token is $w_i \; (w_i \ \mathpunct{:} w_i \in \mathbb{V}, i=1,2,...,s_{len})$, and an individual log sequence is  $l\ =\ (w_1,w_2,...,w_{s_{len}})$. Also mask token is defined $\bar{l}=[MASK]$. The goal of the proposed model $f$ is to determine whether the input log sequence is a normal or abnormal log. The log sequence used during training consists of up to $s_{len}$ number of log keys, and a unique set of log keys is defined as $\mathbb{V}$. The log sequences used for training are all normal logs.

\textbf{Input Representation} The input log sequence is defined as $l$, and the BERT model is used. Accordingly, the input data used in the train and test phases are configured as follows.

\begin{itemize}
\item Train phase : $[CLS],w_1,[MASK],w_3,\cdots ,[SEP]$
\item Test phase : $[CLS],w_1,[MASK],w_3,\cdots ,[SEP]$ * $s_{len}$-times
\end{itemize}

In the train phase, the existing log keys were substituted with the [MASK] token at an arbitrary probability; masking was conducted in this study at 20\%. In the test phase, masking was not performed at an arbitrary probability; however, each log key was replaced with the [MASK] token when one log sequence was given to generate an $s_{len}$ number of data for the test.

\textbf{Objective Function} The objective function used for training is as follows, which is identical to Eq. (2). $m_i$ indicates masking, where $w_i$ is the [MASK] token when $m_i=1$. Furthermore, $H_\theta$ indicates the hidden vector of a transformer encoder. 

\begin{equation} \label{eq:3}
\begin{split}
    & {Max}_\theta\ log\ p(\bar{l}|\hat{l})\\
    & \approx \sum_{i=1}^{s_{len}}{m_i\ log\ p_\theta(w_i|\hat{l})}\\
    & = {\mathlarger{\sum}_{i=1}^{s_{len}}
    {m_i\ log\ \frac{exp(H_\theta({(\hat{l})}_i^\intercal e(w_i))}{\sum_{w^\prime}
    {exp(H_\theta({(\hat{l})}_i^\intercal e(w^\prime))}}p_\theta(w_i|\hat{X})}}
\end{split}
\end{equation}

\subsection{LAnoBERT}
\label{sec:LAnoBERT}

\subsubsection{Overview} 
\label{sec:Overview}

\begin{figure*}[ht]
\centering
\includegraphics[width=0.8\textwidth]{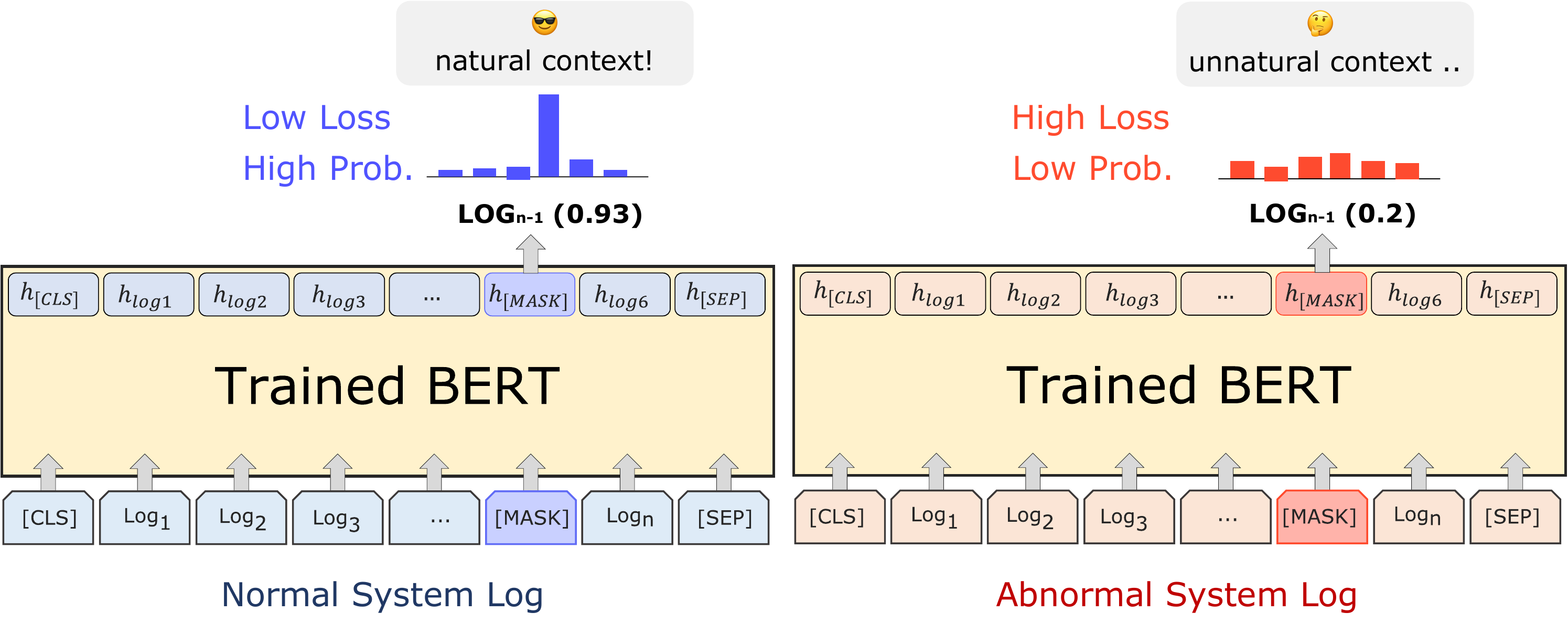}
\caption{Anomaly score distribution difference between normal and abnormal log sequences.}
\label{fig:Anomaly_Score_Distribution}
\end{figure*}

LAnoBERT proposed in this study can be largely divided into the following three parts: preprocessing, model training, and abnormal score computation.

First, the minimum preprocessing of a log sequence was performed in the preprocessing step. Numbers, IPs, and dates are preprocessed, and information loss was minimized using regular expressions. An initialized BERT was used as the model. During the training process, MLM was performed using only normal logs, and masking was randomly performed at 20\%. The NSP objective function was not used in this study when training BERT. Recent studies have pointed out that the NSP objective function interferes with the performance improvement \cite{joshi2020spanbert, DBLP:journals/corr/abs-1901-07291, liu2019roberta, yang2019xlnet}, and it was excluded as it was unnecessary in log anomaly detection. Abnormal scores were calculated from the BERT model, which had been trained using both normal and abnormal logs during test. 

\subsubsection{Preprocessing}
\label{sec:Preprocessing}

\begin{figure}[t]
\centering
\includegraphics[width=\columnwidth]{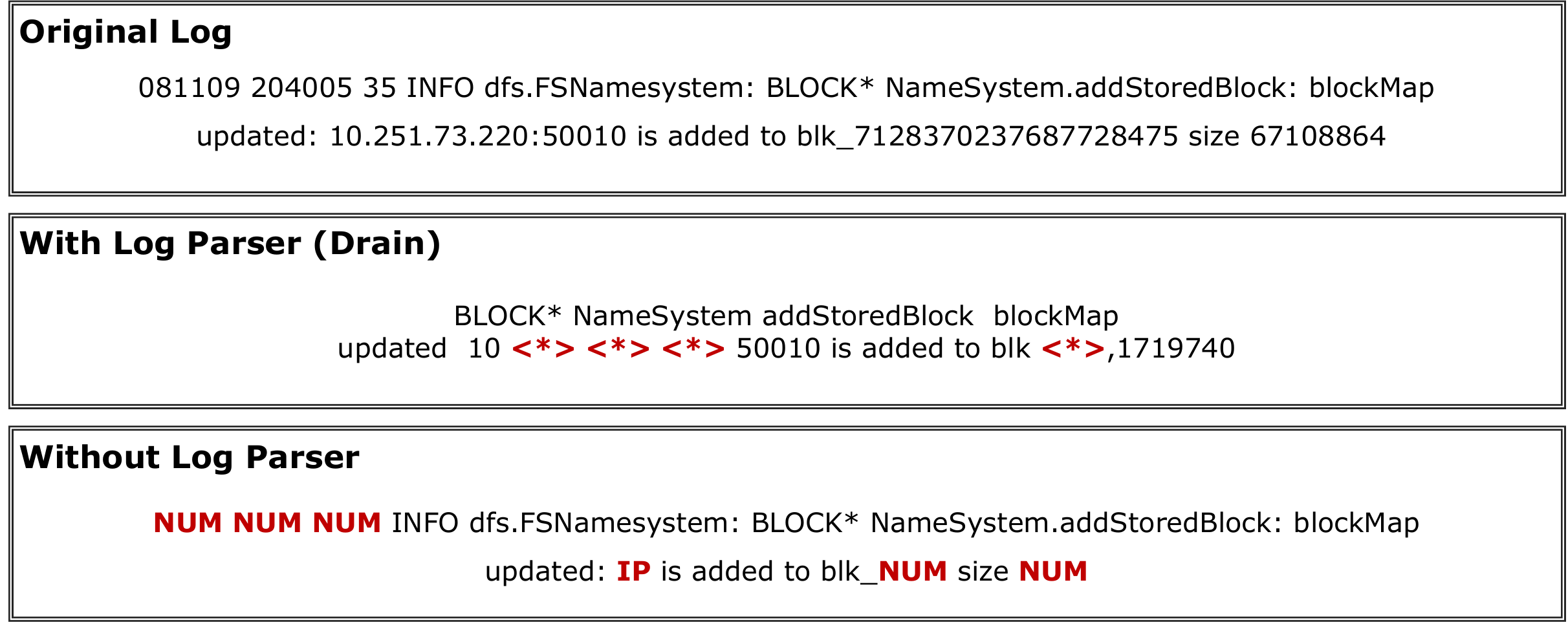}
\caption{Examples of preprocessing system log (HDFS dataset).}
\label{fig:Preprocessing_Systemlog}
\end{figure}

Because this study adopted a log parser-free method, simple preprocessing is conducted using regular expressions. As shown in Figure \ref{fig:Preprocessing_Systemlog}, the original log is highly complicated, unstructured data. When the \textit{Drain} parser is used (with a log parser), the parts defined as a template are excluded and eliminated, whereas certain parts are replaced with \textbf{\textlangle*\textrangle}. Conversely, this study did not use a log parser and instead replaced the data with clear formats such as numbers, dates, and IPs with the words ‘NUM’ or ‘IP’. Preprocessed log sequences were tokenized using the WordPiece \cite{wu2016google} model used in BERT. The tokenizer for the log data was also trained from scratch to ensure that the vocabulary of the log data from each system could be learned. The training was performed only for the normal logs, and the tokenizer created in the training process was used as it was during the test.

\subsubsection{Model}
\label{sec:Model}

The proposed model LAnoBERT executed anomaly detection based on a BERT Masked language model. The most crucial assumption of this study is that \textbf{\textit{``There is a difference between the context of a normal system log and that of an abnormal system log."}} In other words, language models trained only with normal log data are expected to exhibit significant errors and low predictive probability when they encounter the context of abnormal logs during the test. The prediction error defined in this study refers to a cross-entropy loss that occurs between the label information and logit value generated when the model predicts [MASK] as a specific token. Additionally, the predictive probability is defined as a value with the highest probability among the words that can appear in the [MASK] token. When the probability of a predicted word is low, the respective context is considered difficult to find in the normal context and is identified as an anomaly. Therefore, the errors and predictive probability calculated in this process can be utilized in anomaly detection. The core assumption of this study is as shown in Figure \ref{fig:Anomaly_Score_Distribution}. BERT, which is trained only with normal log data, produces low errors and high predictive probability when performing MLM by receiving normal log data as input because the commonly observed logs with a normal pattern are understood as a familiar context. Contrarily, large errors and low predictive probability are produced when performing MLM by receiving abnormal log data as input. The logs are understood as an unfamiliar context if the patterns that are not found in normal logs are input. Previous studies that performed anomaly detection using a transformer defined anomaly detection as a classification problem or adjusted the embedding space using additional objective functions. However, when a log parser is not used in the system log data where normal and abnormal log data are almost identical, a highly complicated format is exhibited and the unsupervised learning-based performance is substantially reduced \cite{nedelkoski2020self}. In this study, therefore, anomaly detection was performed by reflecting the contextuality of the log data. 

\textbf{Train Phase}
Training is performed using the BERT Masked language model for a log sequence that has been tokenized through a tokenizer trained with normal data. Training is initiated from scratch using the initialized BERT, and the same parameters as the BERT-base-uncased are used for the model. The training parameters are almost identical to those of the original BERT \cite{devlin-etal-2019-bert}; the only difference is that the masking probability is set to 0.2. As an unsupervised learning-based anomaly detection model, training is only performed for normal logs.

\textbf{Test Phase}
In the test phase, the trained BERT is used to verify the normal and abnormal logs. Unlike the training phase, all log keys present in the log sequence are applied with masking, and the predictive probability and error value are calculated. At this time, the test is performed by the number of log keys for one log sequence. An example of this process is illustrated in Figure \ref{fig:Test_Phase_Process}.

\subsubsection{Abnormal Score}
\label{sec:Abnormal Score}

\begin{figure*}[t]
\centering
\includegraphics[width=0.7\textwidth]{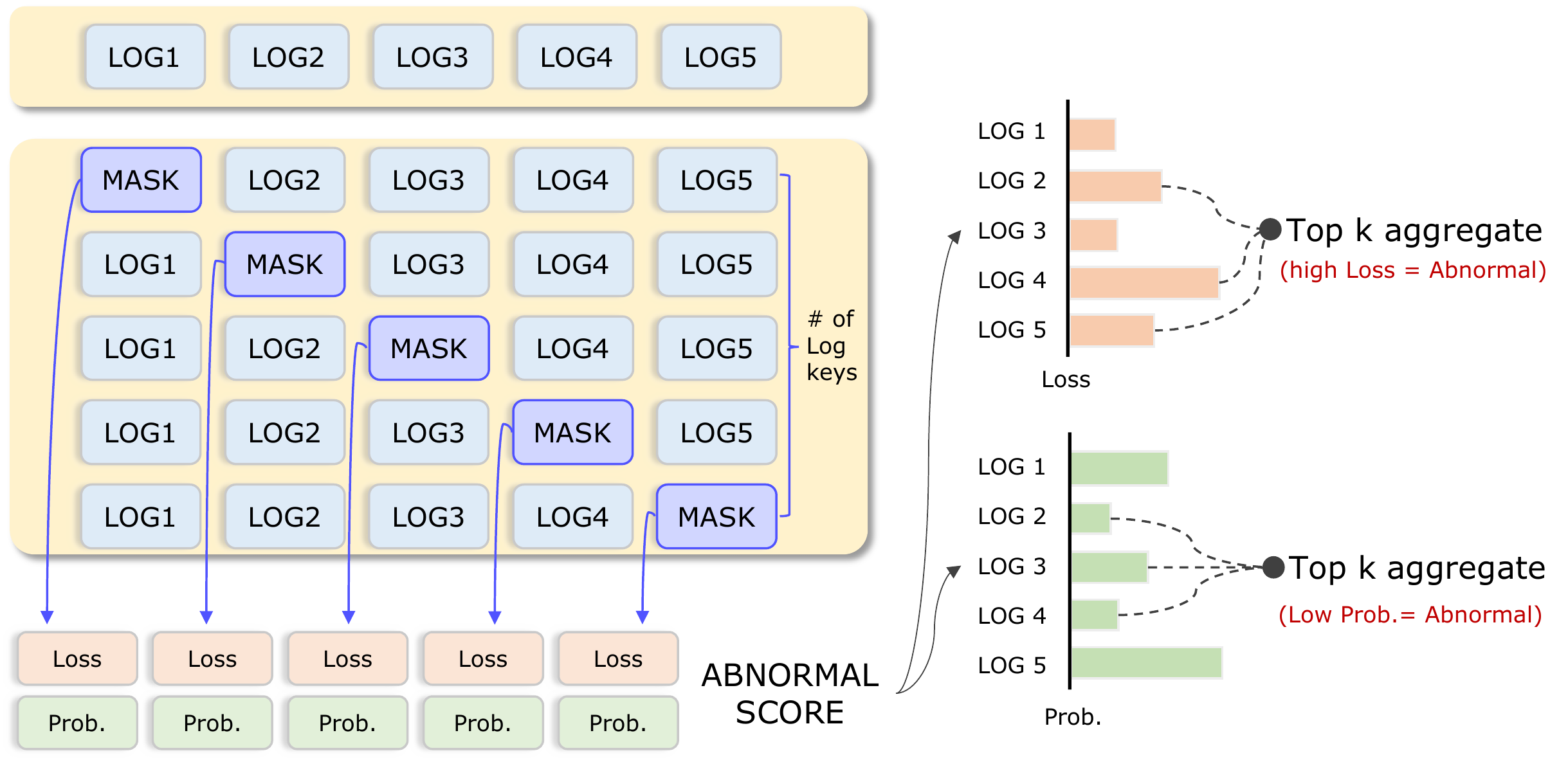}
\caption{Test phase process of LAnoBERT. The proposed model calculates MASK Loss and Prob for each log key to detect anomalies in one log sequence and obtains anomaly scores by aggregating top k values.}
\label{fig:Test_Phase_Process} 
\end{figure*}

As the most important element of anomaly detection, the final abnormal score of one log sequence is defined as the collection of all abnormal scores calculated in the test phase. Owing to the nature of log keys, normal and abnormal logs exhibit almost similar aspects. Therefore, it is not appropriate to use all values calculated for each log key as the abnormal score. If the values of all the log keys are used as the abnormal score, it may cause confusion in recognizing the scores of the abnormal and normal logs. Accordingly, the values calculated from the Top-$k$ number of log keys were used to compute the abnormal score of a given text log in the proposed LAnoBERT. In this study, $k$ is set to 5.

The prediction error proposed in this study and the abnormal score of the predictive probability can be defined as follows. When a log sequence is defined as $l\ =\ (w_1,w_2,\ldots,w_{s_{len}})$, one sequence is generated by the number of log keys to calculate the predictive probability and prediction error. The tokens are then repeatedly replaced with the [MASK] token each; if the $i$-th token is [MASK], the sequence can be defined as ${\hat{l}}_1\ =\ (w_1, w_2, \cdots, w_{i-1}, [MASK], w_{i+1}, w_{s_{len}})$. The number of log sequences used for prediction is identical to the length of the log sequence; thus, the prediction error of the $i$-th log sequence refers to the error value between the logit value calculated in ${\hat{l}}_i$ and the label. Additionally, the predictive probability of the $i$-th log sequence refers to the maximum predictive probability of the word that belongs to the [MASK] position as an answer in ${\hat{l}}_i$. The prediction error and predictive probability, named $error_i$ and $prob_i$ respectively, can be obtained for N number of log sequences. Top $k$ values are selected from the set of calculated prediction errors and predictive probabilities to computer the final abnormal score. The equations for computing the abnormal scores of a test log are shown in Eq. (\ref{eq:4}) and Eq. (\ref{eq:5}). The proposed methodology independently computes the abnormal score with regard to the prediction error and the prediction probability to detect anomalies.
\begin{gather}
{abnormal}_{error}\ =\ \frac{1}{k} \sum_{i \in {Top-k~indices}} error_i, \label{eq:4} \\ 
{abnormal}_{prob}\ =\ \frac{1}{k} \sum_{i \in {Top-k~indices}} prob_i. \label{eq:5} 
\end{gather} 

As mentioned in section \ref{sec:Preprocessing}, LAnoBERT assumed that the abnormal logs would have a large prediction error or a low prediction probability. The abnormal score calculated from each log key is aggregated through the average of the top $k$ values, which becomes the abnormal score of a log sequence. The larger the $abnormal_{error}$, the more likely abnormal a given log sequence, whereas the lower the $abnormal_{prob}$, the more likely abnormal the log sequence.

\begin{figure*}[t]
\centering
\includegraphics[width=0.9\textwidth]{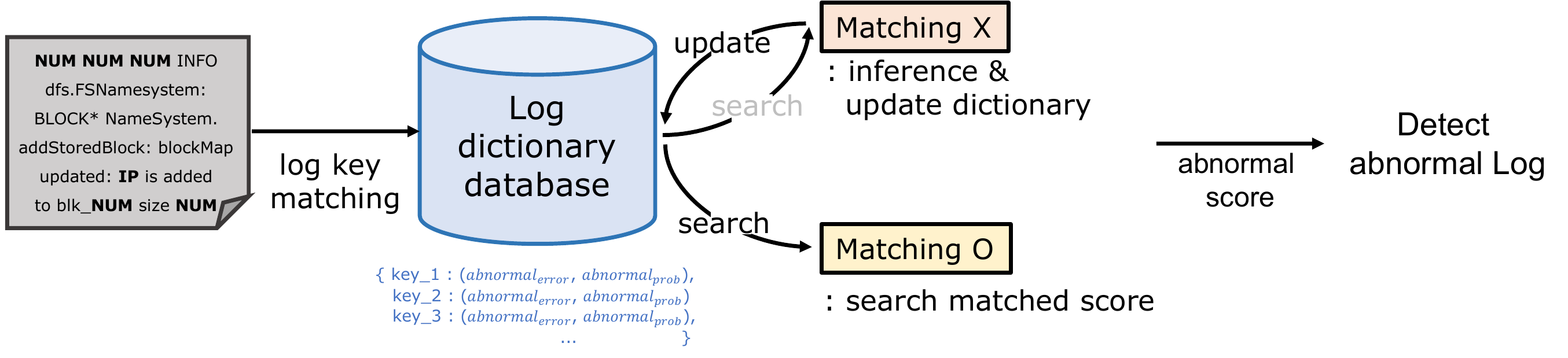}
\caption{Inference process for LAnoBERT}
\label{fig:Effective_Inference}
\end{figure*}

\begin{algorithm}[t!]
    \small
    \DontPrintSemicolon
    \SetAlgoLined
    \SetNoFillComment
    \LinesNotNumbered
    
    \tcp{Definition} 

    \vspace{1.5mm}{
    $KEY$: a set of log sequence keys, $DICT$: the dictionary of the log sequence keys, \\
    $s_{len}$: the sequence length of a log, $d$ : embedding dimension,\\
    $\{l_n\}_{n=1}^N$ : an individual log sequence ($\mathbb{R}^{s_{len} \times d}$), \\
    \text{\textsf{LAnoBERT}} : the proposed model, \text{\textsf{TOPK}} : the top-$k$ aggregation functions \\
    
    \tcp{Initialization}
    
    $KEY \leftarrow \emptyset \,,\,DICT \leftarrow \emptyset\,,\,j=0$ \\
    \SetKwInOut{Input}{Input}
    \SetKwInOut{Output}{Output}
    \Input{$l_n$}
    \Output{$abnormal_{loss}\;,\;abnormal_{prob}$}
    \For{$n=1$ \KwTo $N$}{
    \setstretch{1.3}
        \tcp{log key matching}
        
        \eIf{$l_n$ not in $DICT$}{
            $score_{loss}\leftarrow \emptyset\;,\;score_{prob}\leftarrow \emptyset$ \\
            \For{$i=1$ \KwTo $s_{len}$}{
            $l_{n,i} = [MASK]$ \\
            $loss_j, \; prob_j \leftarrow \text{\textsf{LAnoBERT}}(l_n)$ \\
            $score_{loss} \leftarrow \; score_{loss} \cup \;  \{loss_i\}$ \\
            $score_{prob} \leftarrow \; score_{prob} \cup \;  \{prob_i\}$}
            $abnormal_{loss} \leftarrow \text{\textsf{TOPK}}(score_{loss})$ \\
            $abnormal_{prob} \leftarrow \text{\textsf{TOPK}}(score_{prob})$ \\
            $key_j = l_n$ \\
            $KEY \leftarrow KEY \,  \cup \, \{key_j\}$ \\
            $DICT[key_j] = (abnormal_{loss}\,,\,abnormal_{prob})$ \\
            $j \leftarrow j+1$
            
            \Return{$abnormal_{loss}\,,\,abnormal_{prob}$}
      }{ 
      $abnormal_{loss}\, , \,abnormal_{prob} \leftarrow \, DICT[l_n]$ \\
      \Return{$abnormal_{loss}\,, \,abnormal_{prob}$}
    }}}
    \caption{Inference Process}
\label{algo:Inference algorithm}
\end{algorithm}

However, as shown in Figure \ref{fig:Test_Phase_Process}, when calculating the abnormal score by LAnoBERT for all log sequences existing in the test dataset, the number of required computations becomes the total number of log sequences $\times$ the length of each log sequence. Therefore, if the above method is applied, the computational cost increases and becomes inefficient not sufficient to be applied in an actual system. 

Inspired by the fact that information is accumulated very frequently and there are many duplicates in log data, we propose an efficient inference process by removing repeated computations for duplicated log sequences. Since a masked log sequence for $i^{th}$ token is defined as `$[CLS], w_1, w_2, \cdots, w_{i-1}, [MASK], w_{i+1}, \cdots, w_{s_{len}} , [SEP]$'. We build a log dictionary database with one log sequence as a key value for the inference process. In this database, the dictionary key is defined as a set of $KEY = \{key_0, key_1, key_2, \cdots , key_n\}$. Each key has its corresponding $abnormal_{error}$ and $abnormal_{prob}$ as values: $DICT = \{key_1: (abnormal_{error}, abnormal_{prob}), key_2: (abnormal_{error}, abnormal_{prob}), \cdots, key_j: (abnormal_{error}, abnormal_{prob})\}$. Whenever one log data arrives, the log key matching is performed. If the input key is not matched to any of the existing keys in the current log dictionary, the values for the new key are computed through inference, and then the log dictionary is updated. On the other hand, when the input key is matched to one of the log keys in the current dictionary, the stored values are extracted as the abnormal score without an actual inference process. The following process reduces the unnecessary time required for detecting anomalies by inference of duplicate logs multiple times. Therefore, it is effective because it can be applied in a realistic scenario and is expected to be effective in online anomaly detection settings. An example of this inference process is illustrated in Figure \ref{fig:Effective_Inference} and the algorithm of the entire process is shown in \textsf{Algorithm \ref{algo:Inference algorithm}}.

\section{Experimental Setting}
\label{sec:Experimental Setting}
\subsection{Datasets}
\label{sec:Datasets}

\begin{table}[!t]
\centering
\caption{Number of logs in each dataset used in LAnoBERT}
\label{tab:Dataset}
\resizebox{\columnwidth}{!}{
\setlength{\tabcolsep}{8pt}
\begin{tabular}{cccc} 
\hline
\textbf{Dataset} & \textbf{Type} & \textbf{Train} & \textbf{Test}\\ \hline
\multirow{2}{*}
{\textbf{HDFS}} & normal & 
\begin{tabular}[c]{@{}c@{}}8,712,418\\(446,578 blocks)\end{tabular} & 
\begin{tabular}[c]{@{}c@{}}2,463,201\\(128,483 blocks)\end{tabular}  \\ 
\cline{2-4}  & abnormal & - & \begin{tabular}[c]{@{}c@{}}138,410\\(16,838 blocks)\end{tabular}\\ \hline
\multirow{2}{*}{\textbf{BGL}}  & normal & 3,496,193 & 903,310   \\ 
\cline{2-4} & abnormal & - & 348,460 \\ \hline
\multirow{2}{*}{\textbf{Thunderbird}}  & normal & 166,371,162 & 41,592,791   \\ 
\cline{2-4} & abnormal & - & 3,248,239 \\ \hline
\end{tabular}}
\end{table}

In this study, HDFS \cite{xu2009detecting}, BGL \cite{oliner2007supercomputers} Thunderbird \cite{oliner2007supercomputers} were used as the benchmark log datasets for a fair comparison with previous studies. The three datasets include answer labels, and the generalization performance of the system log anomaly detection model can be verified as the data are deduced from different systems. HDFS, which is the Hadoop Distributed File System, is log data generated from a private cloud environment where one log consists of multiple log sequences. BGL includes data that consist of logs generated by the Blue Gene/L supercomputer, where each individual log sequence is accompanied by a corresponding label indicating either a normal or an abnormal condition.Thunderbird dataset was obtained from the Thunderbird supercomputer system at Sandia National Laboratories (SNL) in Albuquerque. This dataset includes alert and non-alert messages that are identified by alert category tags. Among the three datasets used in this study, the HDFS dataset is considered to have a relatively simple architecture \cite{nedelkoski2020self}, while the Thunderbird dataset has the largest number of log messages. The distribution of normal and abnormal log sequences (or blocks) used in the training and test datasets is presented in Table \ref{tab:Dataset}.

\subsection{Benchmark Methods}
\label{sec:Benchmark Methods}

\begin{figure*}[t!]
\centering
\includegraphics[width=0.9\textwidth]{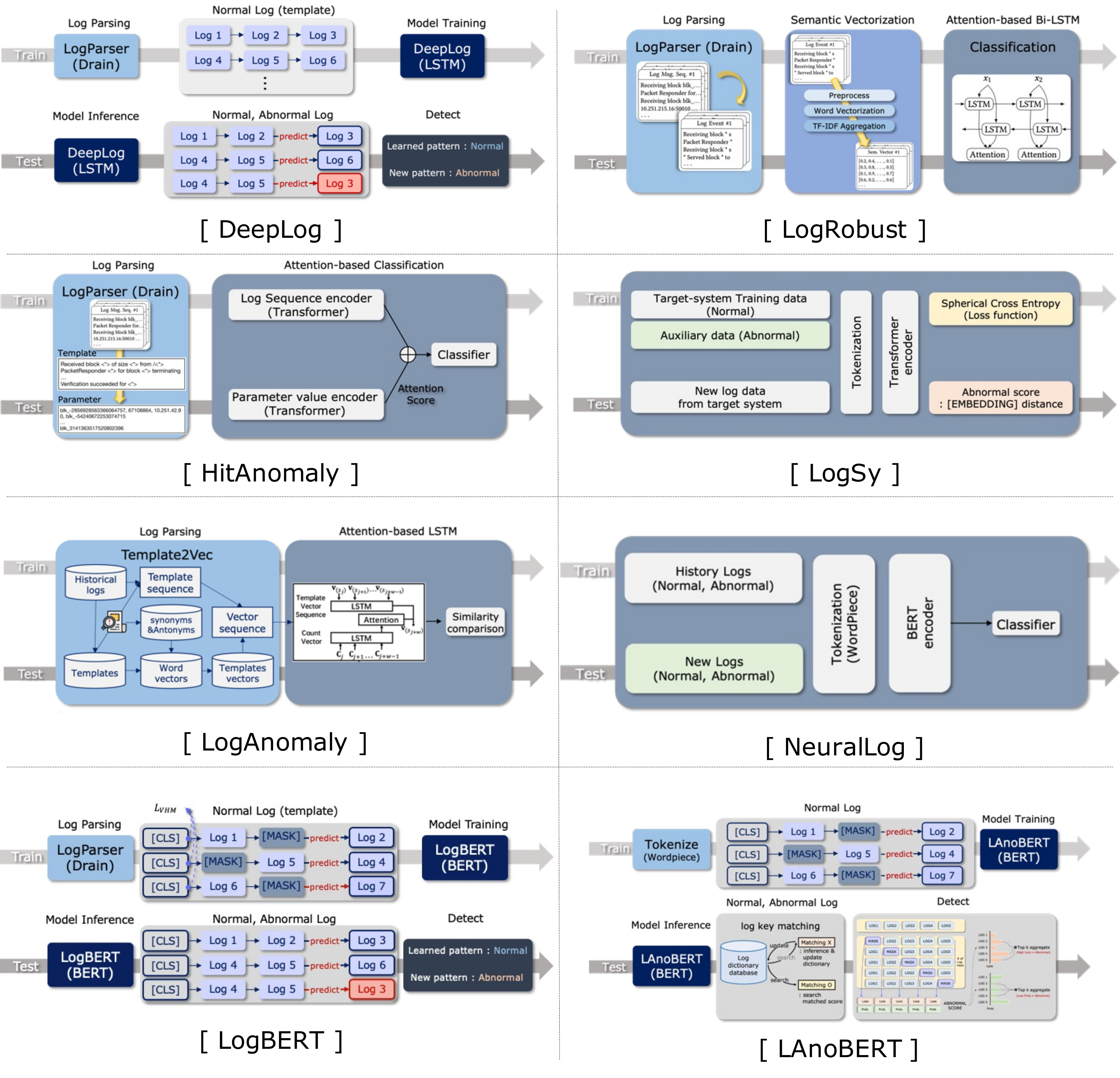}
\caption{Benchmark models and LAnoBERT for log anomaly detection.}
\label{fig:Benchmark_Models}
\end{figure*}

In this section, we present the benchmark models for comparison with LAnoBERT's performance among various log anomaly detection models. The benchmark models were selected based on the usage of a log parser and whether the learning was supervised or unsupervised. The selected models were LogRobust, HitAnomaly, LogSy, Principal Component Analysis (PCA) \cite{xu2009detecting}, One-Class SVM (OCSVM) \cite{10.1162/089976601750264965}, Isolation Forest (iForest) \cite{4781136}, LogCluster \cite{7883294}, DeepLog, LogAnomaly \cite{ijcai2019p658}, and LogBERT. Figure \ref{fig:Benchmark_Models} illustrates the following aspects of the deep learning-based baseline models: 1) structural differences among the models, 2) log data preprocessing method during training and testing, and 3) anomaly detection method.

\begin{itemize}
    \item \textbf{LogRobust} is a supervised learning-based model that utilizes an attention-based bi-LSTM architecture. The model employs a specialized log parser to preprocess log data, generating a TF-IDF and word semantic vector to extract the features of the log data. For training, both normal and abnormal log data are used in solving a classification problem.
    \item \textbf{HitAnomaly} is a supervised learning-based model that employs a transformer architecture. It utilizes a specialized log parser for log data to standardize templates and encode log information as parameters. The model combines two features of normal and abnormal log data with an attention mechanism to classify the data.
    \item \textbf{LogSy} is a supervised learning-based anomaly detection model that utilizes a transformer architecture. It does not require the use of a log parser, as log values are preprocessed using a tokenizer \cite{nedelkoski2020self}. Both normal and abnormal log data are utilized in the model, with data generated from different systems and a distance-based loss function being employed.
    \item \textbf{NeuralLog} is a transformer-based classification model that utilizes a tokenizer. NeuralLog has a similar structure to LogSy, but differs in its utilization of both normal and abnormal data from the target system and a different system during the training process. This approach sets it apart from LogSy, which solves the classification problem by utilizing normal data from the target system and abnormal data generated from a different system. To evaluate the performance of NeuralLog, several popular backbone models, namely BERT, GPT2, and RoBERTa, were utilized. Among these models, BERT achieved the highest performance.
    \item \textbf{PCA} is a linear transformation technique that transforms a set of correlated variables into a set of uncorrelated variables, referred to as principal components. This method builds a counting matrix for log sequence frequency to detect anomalies, then reduces the original counting matrix into a low-dimensional space for the identification of abnormal sequences.
    \item \textbf{OCSVM} is a widely adopted one-class classification model for log anomaly detection \cite{1437839}, utilizing only normal log data. The model is designed to identify the boundary that separates the majority of input data from the remainder, represented as a hyperplane that separates normal data from outliers.
    \item \textbf{iForest} is a tree-based unsupervised learning algorithm that utilizes the isolation of observations that are distinct from the remainder of the input data. The approach employs the formation of an ensemble of decision trees, each of which partitions the data into smaller subsets.
    \item \textbf{LogCluster} is a clustering method for detecting frequently occurring line patterns and abnormal events in textual event logs.
    \item \textbf{DeepLog} is a deep learning-based unsupervised log anomaly detection model based on an LSTM architecture. A specialized log parser is used to generate the input values for the LSTM, and the model predicts the next word. If the next word appears in a trained pattern, it is classified as normal, otherwise, it is considered abnormal.
    \item \textbf{LogAnomaly} is proposed as a solution for detecting anomalies in log streams. The model leverages attention-based LSTM architecture to consider log data as natural language sequences. To extract semantic information, the LogAnomaly model employs the template2vec technique on log templates. This enables the detection of both sequential and quantitative anomalies in log data.
    \item \textbf{LogBERT} is BERT based anomaly detection model that employs MLM and DeepSVDD \cite{pmlr-v80-ruff18a} loss during training. Log data is preprocessed using the log parser, after which the LogBERT model identifies anomalous patterns in the candidate set similar to the DeepLog.
\end{itemize}

\subsection{Evaluation Metrics}
\label{sec:Evaluation Metrics}

\begin{table}[t!]
\caption{Anomaly detection evaluation criteria}
\label{tab:Metric}
\centering
\renewcommand{\arraystretch}{1.0}
\resizebox{\columnwidth}{!}{%
\setlength{\tabcolsep}{15pt}

\begin{tabular}{c|c|c} 
\hline
                  & \textbf{Normal} & \textbf{Abnormal}  \\ 
\hline
\textbf{Normal}   & True Negative   & False Positive     \\ 
\hline
\textbf{Abnormal} & False Negative  & True Positive      \\
\hline
\end{tabular}}
\end{table}

The F1 score which is dependent on the threshold and AUROC, which is independent of the threshold, were used as evaluation metrics in this study. Most studies on anomaly detection use AUROC as the evaluation metric in general, whereas previous studies that approached log anomaly detection as a binary classification problem used the best F1 score to record the performance; hence, both these metrics were used in this study for comparison with previous studies. When the threshold of the models in anomaly detection is determined, the confusion matrix presented in Table \ref{tab:Metric} is generated depending on the actual anomaly case and the anomaly detected by the models. The recall and precision are calculated from the confusion matrix using precision and recall, and the F1 score is calculated based on the harmonic mean of the two metrics.

\begin{gather}
\label{eq: metric}
\text{F1 score} = 2\cdot\frac{\text{precision}\cdot\text{recall}}{\text{precision}+\text{recall}}, \\
\text{precision}=\frac{TP}{TP+FP}, ~~ \text{recall} =\frac{TP}{TP+FN}, \nonumber \\
TP: \text{true positive},  \ FP :  \text{false positive} ,   \  FN: \text{false negative}. \nonumber 
\end{gather}

The F1 score calculated using Eq. (\ref{eq: metric}) is a metric influenced by the threshold of a model and cannot guarantee the reliability of the fundamental performance of an anomaly detection model; hence, AUROC, which is an evaluation metric unaffected by the threshold was also calculated. AUROC calculates the false positive rate (FAR) and true positive rate (TPR) for all the threshold candidates, and then illustrates a receiver operating characteristic curve with the FAR as the x-axis and TPR as the y-axis to calculate the area of the base side. The AUROC value is closer to 1 because the anomaly detection model has a better performance, whereas a random model has a value closer to 0.5. 

The best F1 score threshold cannot be determined in advance in this study because the log anomaly detection experiment is conducted with only normal data for training. Therefore, the best F1 score was calculated using the threshold that represents the best performance theoretically for the test dataset, and the same method was used to calculate the best F1 score of benchmark studies for a fair comparison. Additionally, benchmark studies examined the performance using only the F1 score; however, this study also utilized AUROC, which is an evaluation metric that is not affected by the threshold to examine the performance. In this experiment, the method of determining anomalies differs among the benchmark models. Specifically, DeepLog and LogBERT determine anomalies based on the presence of predicted values in the top $k$ candidate set, while LogRobust and HitAnomaly, being classification models, do not calculate an abnormal score. On the other hand, LogSy, LogCluster, and LogAnomaly define an abnormal score, but the AUROC could not be calculated because no official implementation code was available. Therefore, it is impossible to calculate AUROC for the baseline models for comparison; the F1 score and performance of AUROC of LAnoBERT are presented in Tables \ref{tab:F1_AUROC} and \ref{tab:pretrained_init}. 

\section{Results}
\label{sec:Results}

To verify the excellence of the proposed methodology, this study compared a supervised and an unsupervised learning-based anomaly detection model. Moreover, the use of a parser was recorded for comparison because the performance of the log anomaly detection significantly varies depending on the use of a log parser.

\begin{table*}[t!]
\centering
\caption{F1-score on HDFS, BGL, and Thunderbird. ${}^\dagger$ indicates the performance of benchmark models reported by LogBERT. The highest performance is highlighted in bold and underlined, and the second-best performance is indicated in bold. \textbf{Supervised comparisons (Upper):} the performance of LogRobust, HitAnomaly, LogSy, and NeuralLog are compared, and it is observed that LAnoBERT demonstrates comparable or superior performance to these models, despite the fact that LogRobust, HitAnomaly, LogSy, and NeuralLog allow for the use of abnormal data in their training, whereas LAnoBERT does not. \textbf{Unsupervised comparisons (Lower)}: it is shown that LAnoBERT, which is a log parser-free model, produces strong results compared to other unsupervised models.}
\label{tab:Main_Result}
\resizebox{0.8\textwidth}{!}{%
\setlength{\tabcolsep}{13pt}
\renewcommand{\arraystretch}{1.3}
\begin{tabular}{lcccc}
\hline
                                                                               & \textbf{Log Parser} & \textbf{HDFS}   & \textbf{BGL}    & \textbf{Thunderbird} \\ \hline
\multicolumn{5}{l}{\textit{Supervised}}                                                                                                                         \\ \hline
\textbf{LogRobust}                                                             & O                   & \textbf{0.9700}          & 0.8300          & -                    \\
\textbf{HitAnomaly}                                                            & O                   & \underline{\textbf{0.9800}} & \textbf{0.9200} & -                    \\ 
\textbf{LogSy}                                                                 & X                   & -               & 0.6500          & \underline{\textbf{0.9900}} \\
\textbf{NeuralLog}                                                                 & X                   & \underline{\textbf{0.9800}}               &  \underline{\textbf{0.9800}}          & {\textbf{0.9600}}\\\hline
\multicolumn{5}{l}{\textit{Unsupervised}}                                                                                                                       \\ \hline
\textbf{PCA$^\dagger$}                                                                  & O                   & 0.1112          & 0.1661          & 0.5439               \\
\textbf{iForest$^\dagger$}                                                              & O                   & 0.6049          & 0.3065          & 0.0329               \\
\textbf{OCSVM$^\dagger$}                                                                & O                   & 0.0495          & 0.0196          & 0.2548               \\
\textbf{LogCluster$^\dagger$}                                                           & O                   & 0.5399          & 0.7663          & 0.5961               \\
\textbf{DeepLog$^\dagger$}                                                              & O                   & 0.7734          & 0.8612          & 0.9308               \\
\textbf{LogAnomaly$^\dagger$}                                                           & O                   & 0.5619          & 0.7409          & 0.9273               \\
\textbf{LogBERT}                                                               & O                   & 0.8232          &  \underline{\textbf{0.9083}}          & \textbf{0.9664}               \\ \cline{1-1}
\textbf{\begin{tabular}[c]{@{}l@{}}LAnoBERT\\ (Predictive Loss)\end{tabular}}  & X                   & \textbf{0.9123}          & 0.6932          & 0.5142                    \\
\textbf{\begin{tabular}[c]{@{}l@{}}LAnoBERT\\ (Predictive Prob.)\end{tabular}} & X                   & \underline{\textbf{0.9645}}    & \textbf{0.8749}    & \underline{\textbf{0.9990}}      \\ \hline
\end{tabular}}
\end{table*}

\subsection{Anomaly Detection Performance}
\label{sec:Anomaly Detection Performance}

Table \ref{tab:Main_Result} shows the F1-score for the proposed LAnoBERT model and ten additional models, including both supervised learning-based models (LogSy, LogRobust, and HitAnomaly) and unsupervised learning-based models (PCA, IForest, OCSVM, LogCluster, DeepLog, LogAnomaly, and LogBERT). It is important to note that the performance results for the supervised models were obtained from their respective original studies, whereas the performance of all unsupervised models, except for the proposed LAnoBERT, were obtained from the LogBERT study. As a result, the performance results for LogRobust and HitAnomaly on the Thunderbird dataset, as well as for LogSy on the HDFS dataset, are not reported in this study due to the lack of information in their respective original papers.
The results indicate that the performance of the BGL dataset was inferior compared to the HDFS dataset due to its more complex structure, as previously reported in \citet{huang2020hitanomaly}.

The performance of LogRobust and HitAnomaly, which are based on supervised learning, was observed to be favorable on the HDFS and BGL datasets. Both models underwent preprocessing utilizing the \textit{Drain} parser, and HitAnomaly, which utilized parameters that were not part of the log template, demonstrated strong performance on both datasets. These results indicate that information loss during log parsing can have a significant impact on model training. LogSy, which employed a classification model built with normal and abnormal data obtained from different systems, recorded an F1 score of 0.6500 on the BGL dataset and 0.9900 on the Thunderbird dataset. This highlights the advantage of incorporating a more realistic representation of the system into the model, as compared to LogRobust and HitAnomaly. However, the performance of LogSy on the BGL dataset was lower than expected. These results emphasize the limitations of performing log anomaly detection without log parsing on data from a specific system. NeuralLog demonstrated high performance across three datasets - HDFS, BGL, and Thunderbird, with scores of 0.9800, 0.9800, and 0.9600 respectively. This performance was noteworthy, particularly considering that it didn't utilize a parser, yet still outperformed supervised learning-based models such as LogRobust and HitAnomaly. This outcome can be interpreted as a meaningful result in itself. However, a limitation of NeuralLog is its reliance on both normal and abnormal logs during the learning process, which could make it less suitable for real-world scenarios. This point can be identified as a potential shortcoming of the model.

In the unsupervised learning models comparison, PCA, iForest, OCSVM, and LogCluster showed lower performance compared to DeepLog, LogAnomaly, and LogBERT which utilized deep learning techniques. Specifically, DeepLog outperformed LogAnomaly, demonstrating the effectiveness of its "top $g$ candidate" logic. LogBERT demonstrated remarkable performance with F1 scores of 0.8232 in HDFS, 0.9083 in BGL, and 0.9664 in Thunderbird, with especially strong results in BGL. These results suggest that the BERT-based LogBERT model effectively captures rich semantics by understanding context-specific information to log data. Furthermore, incorporating MLKP and VHM tasks has been observed to improve the model's ability to detect anomalies.

In Section \ref{sec:Abnormal Score}, it was highlighted that BERT's MLM allows for the calculation of both mask loss and probability. In order to evaluate the performance of the proposed LAnoBERT, two abnormal scores were generated using mask loss and probability. The results showed that the predictive loss score led to a performance of 0.9123 in HDFS, 0.6932 in BGL, and 0.5142 in Thunderbird. It was observed that HDFS, with its shorter log sequence length and fewer unique log keys, displayed acceptable detection performance. Conversely, BGL and Thunderbird, characterized by longer log sequence lengths and more complex structures, showed relatively lower performance than other deep learning-based unsupervised models. This can be attributed to the fact that the mask loss calculates the accuracy of word-by-word predictions between the original and predicted log keys, resulting in low loss values only when the log keys are predicted in the correct order. For example, if the ground truth log key is \texttt{`A-B-C-D-E'} and the model predicts \texttt{`B-C-D-E-F'}, the loss value would be high due to the incorrect prediction of all tokens. However, from the perspective of the log sequence, the confidence in the ordered prediction of the keys \texttt{`B-C-D-E'} must be considered when evaluating an abnormal score. The previously discussed example highlights the limitations of using the predictive loss score on long and complex data.

When the predictive probability was used as the abnormal score, LAnoBERT demonstrated superior performance to all the other models, with HDFS scoring 0.9645, BGL scoring 0.8749, and Thunderbird scoring 0.9990, with the exception of BGL. The results showed that the abnormal score based on the mask probability proposed by LAnoBERT was a critical factor in performance improvement. These results highlight the effectiveness of LAnoBERT, an unsupervised learning-based method, compared to the parser-based supervised learning methodologies LogRobust, HitAnomaly, and LogSy. Despite not using a parser during training, LAnoBERT achieved higher performance than LogRobust and LogSy, while being only 0.0451 lower than HitAnomaly in BGL. This demonstrates the significance of considering context and pre-training of MLM in the design of LAnoBERT for log anomaly detection. Additionally, using predictive probability allows for the detection of cases with unseen normal log data more accurately compared to using predictive loss, making LAnoBERT more practical and useful in real-world applications than benchmark models.

\begin{figure*}[t!]
\centering
\includegraphics[width=1\textwidth]{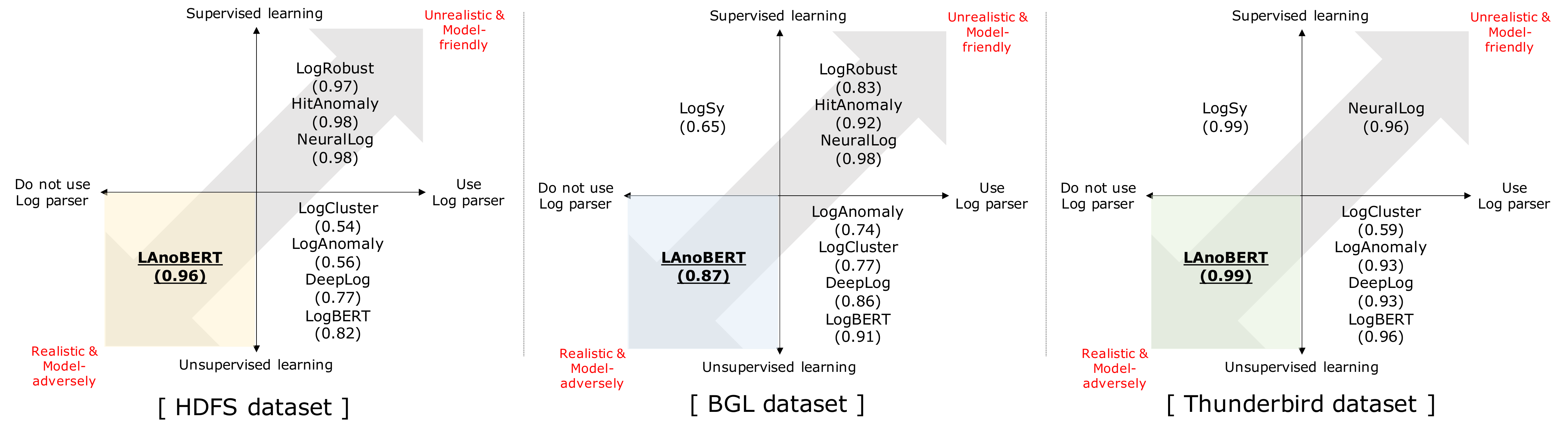}
\caption{Comparison with LAnoBERT and four benchmark models in realistic scenarios.}
\label{fig:Realistic_Scenarios}
\end{figure*}

Furthermore, it is critical to perform log anomaly detection on actual systems. Since logs are collected in real-time, the majority of log data is comprised of normal logs. As a result, conducting anomaly detection based on binary classification using normal and abnormal log data poses a significant challenge for its practical implementation. Figure \ref{fig:Realistic_Scenarios} illustrates the selected benchmark models of supervised and unsupervised learning, with and without the use of a parser. The first quadrant, which represents the parser-involved supervised case, represents the easiest scenario to ensure the performance of a model, but it is also the most unrealistic. On the other hand, the parser-free unsupervised case in the third quadrant is the most realistic scenario but also the most challenging in terms of ensuring the performance of a model. Despite these challenges, the proposed LAnoBERT model in this study showed comparable performance (BGL) to benchmark models under the easiest assumption and even better performance (HDFS, Thunderbird) than some benchmark models in the most difficult scenario. Hence, the LAnoBERT model proposed in this study appears to be a practical model that can be applied in real-world systems.

\begin{table*}[t!]
\caption{Performance of our model (F1 Score, AUROC)}
\label{tab:F1_AUROC}
\centering
\resizebox{0.8\textwidth}{!}{%
\setlength{\tabcolsep}{8pt}
\renewcommand{\arraystretch}{1.2}
\begin{tabular}{lcclclcl}
\hline
\multicolumn{1}{l|}{\multirow{2}{*}{}}                                                              & \multicolumn{1}{c|}{\multirow{2}{*}{\textbf{Log Parser}}} & \multicolumn{2}{c|}{\textbf{HDFS}}                          & \multicolumn{2}{c|}{\textbf{BGL}}                           & \multicolumn{2}{c}{\textbf{Thunderbird}}                            \\ \cline{3-8} 
\multicolumn{1}{l|}{}                                                                               & \multicolumn{1}{c|}{}                                     & \multicolumn{1}{l|}{F1 Score} & \multicolumn{1}{l|}{AUROC}  & \multicolumn{1}{l|}{F1 Score} & \multicolumn{1}{l|}{AUROC}  & \multicolumn{1}{l|}{F1 Score} & AUROC                               \\ \hline
\multicolumn{8}{l}{\textit{Unsupervised}}                                                                                                                                                                                                                                                                                                                         \\ \hline
\multicolumn{1}{l|}{\textbf{\begin{tabular}[c]{@{}l@{}}LAnoBERT\\ (Predictive Prob.)\end{tabular}}} & \multicolumn{1}{c|}{X}                                    & \multicolumn{1}{c|}{0.9645}   & \multicolumn{1}{c|}{0.9901} & \multicolumn{1}{c|}{0.8749}   & \multicolumn{1}{c|}{0.9721} & \multicolumn{1}{c|}{0.9990}   & \multicolumn{1}{c}{0.9520} \\ \hline
\end{tabular}}
\end{table*}

As explained earlier, supervised learning-based models are still proposed for system log anomaly detection. The proposed model achieved better performance than LogSy, which is a supervised learning-based model, and the LogRobust model, which uses a parser and is a supervised learning-based model. LAnoBERT recorded excellent detection performance than certain supervised learning-based baseline models in the most unfair comparison environment. In other words, LAnoBERT is a model trained in the most realistic yet disadvantageous environment and is a robust model exhibiting more outstanding or similar performance compared to other baseline models in the most unrealistic and advantageous environment.

Table \ref{tab:F1_AUROC} lists the F1 score and AUROC performance of LAnoBERT. The performance of benchmark models was not measured using AUROC, which is a frequently used evaluation metric in anomaly detection. benchmark models may score highly for the best F1 score to exhibit the best performance; however, they are limited in identifying whether the model has high reliability regardless of the threshold. Therefore, the performances of the two evaluation metrics were determined for LAnoBERT, and the results showed that the F1 score is similar to that of the other models evaluated in a relatively more advantageous environment. By contrast, a high AUROC value closer to 1, which indicates that a model is most idealistic, was obtained. Consequently, even if threshold-dependent detection performance metrics other than the F1 score are used, LAnoBERT can be regarded as a highly reliable system log anomaly detection model with remarkably outstanding performance.

\subsection{Performance according to the BERT learning method}
\label{sec:Performance according to the BERT learning method}

\begin{table*}[t!]
\caption{Performance of initialized LAnoBERT and pre-trained LAnoBERT}
\label{tab:pretrained_init}
\centering
\resizebox{0.8\textwidth}{!}{%
\setlength{\tabcolsep}{5pt}
\renewcommand{\arraystretch}{1.2}
\begin{tabular}{lcccccccc}
\hline
\multicolumn{1}{l|}{\multirow{2}{*}{}}                                                              & \multicolumn{1}{c|}{\multirow{2}{*}{\textbf{Log Parser}}} & \multicolumn{1}{c|}{\multirow{2}{*}{\textbf{Training}}} & \multicolumn{2}{c|}{\textbf{HDFS}}                                          & \multicolumn{2}{c|}{\textbf{BGL}}                                           & \multicolumn{2}{c}{\textbf{Thunderbird}}                  \\ \cline{4-9} 
\multicolumn{1}{l|}{}                                                                               & \multicolumn{1}{c|}{}                                     & \multicolumn{1}{c|}{}                                   & \multicolumn{1}{l|}{F1 Score}        & \multicolumn{1}{l|}{AUROC}           & \multicolumn{1}{l|}{F1 Score}        & \multicolumn{1}{l|}{AUROC}           & \multicolumn{1}{l|}{F1 Score} & \multicolumn{1}{l}{AUROC} \\ \hline
\multicolumn{9}{l}{\textit{Unsupervised}}                                                                                                                                                                                                                                                                                                                                                                                                         \\ \hline
\multicolumn{1}{l|}{\textbf{\begin{tabular}[c]{@{}l@{}}LAnoBERT\\ (Predictive Prob.)\end{tabular}}} & \multicolumn{1}{c|}{X}                                    & \multicolumn{1}{c|}{Initialized}                        & \multicolumn{1}{c|}{\textbf{0.9645}} & \multicolumn{1}{c|}{\textbf{0.9901}} & \multicolumn{1}{c|}{0.8749}          & \multicolumn{1}{c|}{0.9721}          & \multicolumn{1}{c|}{\textbf{0.9990}}   & \textbf{0.9520}           \\
\multicolumn{1}{l|}{\textbf{\begin{tabular}[c]{@{}l@{}}LAnoBERT\\ (Predictive Prob.)\end{tabular}}} & \multicolumn{1}{c|}{X}                                    & \multicolumn{1}{c|}{pre-trained}                         & \multicolumn{1}{c|}{0.9304}          & \multicolumn{1}{c|}{0.9659}          & \multicolumn{1}{c|}{\textbf{0.9020}} & \multicolumn{1}{c|}{\textbf{0.9912}} & \multicolumn{1}{c|}{0.8954}         & 0.3505                 \\ \hline
\end{tabular}}
\end{table*}

BERT includes models pre-trained with natural language, and the pre-trained model typically resulted in an excellent performance in various natural language processing tasks. Therefore, a comparative experiment was conducted for LAnoBERT, which was pre-trained with natural language. The proposed LAnoBERT was trained to utilize an initialized BERT model. In order to investigate the impact of the pre-training model and provide a practical alternative for real-world log anomaly detection, we conducted an additional experiment in which the pre-trained BERT model with natural language is employed instead of training BERT from scratch. The results of this experiment are documented in Table \ref{tab:pretrained_init}.

When the BERT model pre-trained with natural language was used, MLM was additionally performed after importing the pre-trained model. Pre-training has already been performed with massive natural language data, and thus, it can be interpreted that task adaptive pre-training \cite{gururangan-etal-2020-dont} was conducted with the log data. Table \ref{tab:pretrained_init} presents the result of training 6,000 steps with a batch size of 15 per 2080 ti for a total of two 2080 ti’s. 

When the BERT model pre-trained with natural language was used, the F1 score in the BGL data was 0.9020, which was improved by 0.0271 compared to the model trained from scratch; by contrast, the F1 score in the HDFS data was 0.9304, which was decreased by 0.0341 compared to the model trained from scratch. These results indicate that the HDFS data consisting of a very simple log structure have degraded performance when a model that has learned the context of natural language is used. The number of vocabularies in the HDFS dataset after preprocessing is only 200, which is very few for expressing the context of natural language; hence, the pre-training presumably had a negative effect on anomaly detection. Conversely, BGL data with a relatively more complicated structure has a total of 1,000 vocabularies after preprocessing. This supports the fact that the BGL dataset is a more complicated dataset than HDFS, and there are cases where natural language is included in the log data with this number of vocabularies. Therefore, if appropriate training is performed additionally for a model that has learned the context of natural language, the anomaly detection performance can be improved compared to other models that have not been pre-trained. The experimental results in Table \ref{tab:pretrained_init} show that the log data containing natural language can have a similar form as human language and that pre-trained BERT can be effectively applied. In conclusion, the results demonstrate that incorporating the LAnoBERT framework with a pre-trained BERT model is a viable alternative.

\section{Conclusion}
\label{sec:Conclusion}

This study proposed LAnoBERT, which is an unsupervised learning-based system log anomaly detection model where a parser is not used. The proposed LAnoBERT learned the context of normal log data using MLM, and abnormal logs were detected based on the prediction error and predictive probability during the test. In terms of the nature of the system log, normal and abnormal data have similar characteristics; thus, a new score calculation method is proposed for defining the abnormal score based on the top-k predictive probability. The proposed model exhibited the best performance compared to the unsupervised models, and superior or similar performance compared to supervised learning-based models. In addition, the efficient inference process proposed in this study is expected to work well in an actual system. Although the performances of benchmark models are heavily dependent on the use of log parser, our proposed LAnoBERT can be a robust and parser-independent log anomaly detection model.

The proposed LAnoBERT framework exhibits promising results in log anomaly detection, however, there are limitations that need to be addressed in future research. Firstly, LAnoBERT requires individual training for each log dataset. A unified framework, as outlined in UniAD \cite{you2022a}, is needed to cater to diverse log structures in different systems like distributed systems, supercomputers, and server applications. Secondly, LAnoBERT's Transformer-based architecture incurs higher computational costs compared to RNN-based models due to its self-attention layer ($O(n^2 \cdot d)$ complexity) versus the recurrent layer of RNN ($O(n \cdot d^2)$ complexity). To resolve the computational inefficiency, incorporating recent parameter-efficient learning methods such as LoRA \cite{hu2022lora} and Adapter \cite{pmlr-v97-houlsby19a} is crucial in developing a real-time log anomaly detection model. Finally, in this study, only minimal preprocessing was performed using regular expressions and tokenization using Wordpiece tokenizer. The Log Parser-free methodology can be improved by templating log sequences into the natural language via prompt tuning \cite{NEURIPS2020_1457c0d6, lester-etal-2021-power} which could enable anomaly detection with a pre-trained tokenizer and language model, without the need for further preprocessing or training.


\section*{Acknowledgements}
This work was supported by the National Research Foundation of Korea (NRF) grant funded by the Korea government (MSIT) (NRF-2022R1A2C2005455). This work was also supported by Institute of Information \& communications Technology Planning \& Evaluation (IITP) grant funded by the Korea government (MSIT) (No. 2021-0-00471, Development of Autonomous Control Technology for Error-Free Information Infrastructure Based on Modeling \& Optimization).

\bibliography{main}

\begin{thebibliography}{48}
\providecommand{\natexlab}[1]{#1}
\providecommand{\url}[1]{\texttt{#1}}
\expandafter\ifx\csname urlstyle\endcsname\relax
  \providecommand{\doi}[1]{doi: #1}\else
  \providecommand{\doi}{doi: \begingroup \urlstyle{rm}\Url}\fi

\bibitem[Brown et~al.(2018)Brown, Tuor, Hutchinson, and
  Nichols]{10.1145/3217871.3217872}
Brown, A., Tuor, A., Hutchinson, B., and Nichols, N.
\newblock Recurrent neural network attention mechanisms for interpretable
  system log anomaly detection.
\newblock In \emph{Proceedings of the First Workshop on Machine Learning for
  Computing Systems}, MLCS'18, pp.\ ~8, New York, NY, USA, 2018. Association
  for Computing Machinery.
\newblock ISBN 9781450358651.
\newblock \doi{10.1145/3217871.3217872}.
\newblock URL \url{https://doi.org/10.1145/3217871.3217872}.

\bibitem[Brown et~al.(2020)Brown, Mann, Ryder, Subbiah, Kaplan, Dhariwal,
  Neelakantan, Shyam, Sastry, Askell, Agarwal, Herbert-Voss, Krueger, Henighan,
  Child, Ramesh, Ziegler, Wu, Winter, Hesse, Chen, Sigler, Litwin, Gray, Chess,
  Clark, Berner, McCandlish, Radford, Sutskever, and
  Amodei]{NEURIPS2020_1457c0d6}
Brown, T., Mann, B., Ryder, N., Subbiah, M., Kaplan, J.~D., Dhariwal, P.,
  Neelakantan, A., Shyam, P., Sastry, G., Askell, A., Agarwal, S.,
  Herbert-Voss, A., Krueger, G., Henighan, T., Child, R., Ramesh, A., Ziegler,
  D., Wu, J., Winter, C., Hesse, C., Chen, M., Sigler, E., Litwin, M., Gray,
  S., Chess, B., Clark, J., Berner, C., McCandlish, S., Radford, A., Sutskever,
  I., and Amodei, D.
\newblock Language models are few-shot learners.
\newblock In Larochelle, H., Ranzato, M., Hadsell, R., Balcan, M., and Lin, H.
  (eds.), \emph{Advances in Neural Information Processing Systems}, volume~33,
  pp.\  1877--1901. Curran Associates, Inc., 2020.
\newblock URL
  \url{https://proceedings.neurips.cc/paper/2020/file/1457c0d6bfcb4967418bfb8ac142f64a-Paper.pdf}.

\bibitem[Cinque et~al.(2013)Cinque, Cotroneo, and Pecchia]{6320555}
Cinque, M., Cotroneo, D., and Pecchia, A.
\newblock Event logs for the analysis of software failures: A rule-based
  approach.
\newblock \emph{IEEE Transactions on Software Engineering}, 39\penalty0
  (6):\penalty0 806--821, 2013.
\newblock \doi{10.1109/TSE.2012.67}.

\bibitem[Clark et~al.(2019)Clark, Khandelwal, Levy, and
  Manning]{clark-etal-2019-bert}
Clark, K., Khandelwal, U., Levy, O., and Manning, C.~D.
\newblock What does {BERT} look at? an analysis of {BERT}{'}s attention.
\newblock In \emph{Proceedings of the 2019 ACL Workshop BlackboxNLP: Analyzing
  and Interpreting Neural Networks for NLP}, pp.\  276--286, Florence, Italy,
  August 2019. Association for Computational Linguistics.
\newblock \doi{10.18653/v1/W19-4828}.
\newblock URL \url{https://aclanthology.org/W19-4828}.

\bibitem[Devlin et~al.(2019)Devlin, Chang, Lee, and
  Toutanova]{devlin-etal-2019-bert}
Devlin, J., Chang, M.-W., Lee, K., and Toutanova, K.
\newblock {BERT}: Pre-training of deep bidirectional transformers for language
  understanding.
\newblock In \emph{Proceedings of the 2019 Conference of the North {A}merican
  Chapter of the Association for Computational Linguistics: Human Language
  Technologies, Volume 1 (Long and Short Papers)}, pp.\  4171--4186,
  Minneapolis, Minnesota, June 2019. Association for Computational Linguistics.
\newblock \doi{10.18653/v1/N19-1423}.
\newblock URL \url{https://aclanthology.org/N19-1423}.

\bibitem[Du \& Li(2016)Du and Li]{7837916}
Du, M. and Li, F.
\newblock Spell: Streaming parsing of system event logs.
\newblock In \emph{2016 IEEE 16th International Conference on Data Mining
  (ICDM)}, pp.\  859--864, 2016.
\newblock \doi{10.1109/ICDM.2016.0103}.

\bibitem[Du et~al.(2017)Du, Li, Zheng, and Srikumar]{10.1145/3133956.3134015}
Du, M., Li, F., Zheng, G., and Srikumar, V.
\newblock Deeplog: Anomaly detection and diagnosis from system logs through
  deep learning.
\newblock In \emph{Proceedings of the 2017 ACM SIGSAC Conference on Computer
  and Communications Security}, CCS '17, pp.\  1285–1298, New York, NY, USA,
  2017. Association for Computing Machinery.
\newblock ISBN 9781450349468.
\newblock \doi{10.1145/3133956.3134015}.
\newblock URL \url{https://doi.org/10.1145/3133956.3134015}.

\bibitem[Elnaggar et~al.(2021)Elnaggar, Heinzinger, Dallago, Rehawi, Yu, Jones,
  Gibbs, Feher, Angerer, Steinegger, Bhowmik, and Rost]{9477085}
Elnaggar, A., Heinzinger, M., Dallago, C., Rehawi, G., Yu, W., Jones, L.,
  Gibbs, T., Feher, T., Angerer, C., Steinegger, M., Bhowmik, D., and Rost, B.
\newblock Prottrans: Towards cracking the language of lifes code through
  self-supervised deep learning and high performance computing.
\newblock \emph{IEEE Transactions on Pattern Analysis and Machine
  Intelligence}, pp.\  1--1, 2021.
\newblock \doi{10.1109/TPAMI.2021.3095381}.

\bibitem[Guo et~al.(2021)Guo, Yuan, and Wu]{9534113}
Guo, H., Yuan, S., and Wu, X.
\newblock Logbert: Log anomaly detection via bert.
\newblock In \emph{2021 International Joint Conference on Neural Networks
  (IJCNN)}, pp.\  1--8, 2021.
\newblock \doi{10.1109/IJCNN52387.2021.9534113}.

\bibitem[Gururangan et~al.(2020)Gururangan, Marasovi{\'c}, Swayamdipta, Lo,
  Beltagy, Downey, and Smith]{gururangan-etal-2020-dont}
Gururangan, S., Marasovi{\'c}, A., Swayamdipta, S., Lo, K., Beltagy, I.,
  Downey, D., and Smith, N.~A.
\newblock Don{'}t stop pretraining: Adapt language models to domains and tasks.
\newblock In \emph{Proceedings of the 58th Annual Meeting of the Association
  for Computational Linguistics}, pp.\  8342--8360, Online, July 2020.
  Association for Computational Linguistics.
\newblock \doi{10.18653/v1/2020.acl-main.740}.
\newblock URL \url{https://aclanthology.org/2020.acl-main.740}.

\bibitem[Hansen \& Atkins(1993)Hansen and Atkins]{hansen1993automated}
Hansen, S.~E. and Atkins, E.~T.
\newblock Automated system monitoring and notification with swatch.
\newblock In \emph{LISA}, volume~93, pp.\  145--152, 1993.

\bibitem[He et~al.(2017)He, Zhu, Zheng, and Lyu]{He2017DrainAO}
He, P., Zhu, J., Zheng, Z., and Lyu, M.~R.
\newblock Drain: An online log parsing approach with fixed depth tree.
\newblock \emph{2017 IEEE International Conference on Web Services (ICWS)},
  pp.\  33--40, 2017.

\bibitem[Houlsby et~al.(2019)Houlsby, Giurgiu, Jastrzebski, Morrone,
  De~Laroussilhe, Gesmundo, Attariyan, and Gelly]{pmlr-v97-houlsby19a}
Houlsby, N., Giurgiu, A., Jastrzebski, S., Morrone, B., De~Laroussilhe, Q.,
  Gesmundo, A., Attariyan, M., and Gelly, S.
\newblock Parameter-efficient transfer learning for {NLP}.
\newblock In Chaudhuri, K. and Salakhutdinov, R. (eds.), \emph{Proceedings of
  the 36th International Conference on Machine Learning}, volume~97 of
  \emph{Proceedings of Machine Learning Research}, pp.\  2790--2799. PMLR,
  09--15 Jun 2019.
\newblock URL \url{https://proceedings.mlr.press/v97/houlsby19a.html}.

\bibitem[Hu et~al.(2022)Hu, yelong shen, Wallis, Allen-Zhu, Li, Wang, Wang, and
  Chen]{hu2022lora}
Hu, E.~J., yelong shen, Wallis, P., Allen-Zhu, Z., Li, Y., Wang, S., Wang, L.,
  and Chen, W.
\newblock Lo{RA}: Low-rank adaptation of large language models.
\newblock In \emph{International Conference on Learning Representations}, 2022.
\newblock URL \url{https://openreview.net/forum?id=nZeVKeeFYf9}.

\bibitem[Huang et~al.(2020)Huang, Liu, Fung, He, Zhao, Yang, and
  Luan]{huang2020hitanomaly}
Huang, S., Liu, Y., Fung, C., He, R., Zhao, Y., Yang, H., and Luan, Z.
\newblock Hitanomaly: Hierarchical transformers for anomaly detection in system
  log.
\newblock \emph{IEEE Transactions on Network and Service Management},
  17\penalty0 (4):\penalty0 2064--2076, 2020.

\bibitem[Jawahar et~al.(2019)Jawahar, Sagot, and
  Seddah]{jawahar-etal-2019-bert}
Jawahar, G., Sagot, B., and Seddah, D.
\newblock What does {BERT} learn about the structure of language?
\newblock In \emph{Proceedings of the 57th Annual Meeting of the Association
  for Computational Linguistics}, pp.\  3651--3657, Florence, Italy, July 2019.
  Association for Computational Linguistics.
\newblock \doi{10.18653/v1/P19-1356}.
\newblock URL \url{https://aclanthology.org/P19-1356}.

\bibitem[Joshi et~al.(2020)Joshi, Chen, Liu, Weld, Zettlemoyer, and
  Levy]{joshi2020spanbert}
Joshi, M., Chen, D., Liu, Y., Weld, D.~S., Zettlemoyer, L., and Levy, O.
\newblock Spanbert: Improving pre-training by representing and predicting
  spans.
\newblock \emph{Transactions of the Association for Computational Linguistics},
  8:\penalty0 64--77, 2020.

\bibitem[Kim et~al.(2016)Kim, Yi, Lee, Paek, and
  Yoon]{DBLP:journals/corr/KimYLPY16}
Kim, G., Yi, H., Lee, J., Paek, Y., and Yoon, S.
\newblock Lstm-based system-call language modeling and robust ensemble method
  for designing host-based intrusion detection systems.
\newblock \emph{CoRR}, abs/1611.01726, 2016.
\newblock URL \url{http://arxiv.org/abs/1611.01726}.

\bibitem[Lample \& Conneau(2019)Lample and
  Conneau]{DBLP:journals/corr/abs-1901-07291}
Lample, G. and Conneau, A.
\newblock Cross-lingual language model pretraining.
\newblock \emph{CoRR}, abs/1901.07291, 2019.
\newblock URL \url{http://arxiv.org/abs/1901.07291}.

\bibitem[Le \& Zhang(2021)Le and Zhang]{le2021log}
Le, V.-H. and Zhang, H.
\newblock Log-based anomaly detection without log parsing.
\newblock In \emph{2021 36th IEEE/ACM International Conference on Automated
  Software Engineering (ASE)}, pp.\  492--504. IEEE, 2021.

\bibitem[Lester et~al.(2021)Lester, Al-Rfou, and
  Constant]{lester-etal-2021-power}
Lester, B., Al-Rfou, R., and Constant, N.
\newblock The power of scale for parameter-efficient prompt tuning.
\newblock In \emph{Proceedings of the 2021 Conference on Empirical Methods in
  Natural Language Processing}, pp.\  3045--3059, Online and Punta Cana,
  Dominican Republic, November 2021. Association for Computational Linguistics.
\newblock \doi{10.18653/v1/2021.emnlp-main.243}.
\newblock URL \url{https://aclanthology.org/2021.emnlp-main.243}.

\bibitem[Lin et~al.(2016)Lin, Zhang, Lou, Zhang, and Chen]{7883294}
Lin, Q., Zhang, H., Lou, J.-G., Zhang, Y., and Chen, X.
\newblock Log clustering based problem identification for online service
  systems.
\newblock In \emph{2016 IEEE/ACM 38th International Conference on Software
  Engineering Companion (ICSE-C)}, pp.\  102--111, 2016.

\bibitem[Liu et~al.(2008)Liu, Ting, and Zhou]{4781136}
Liu, F.~T., Ting, K.~M., and Zhou, Z.-H.
\newblock Isolation forest.
\newblock In \emph{2008 Eighth IEEE International Conference on Data Mining},
  pp.\  413--422, 2008.
\newblock \doi{10.1109/ICDM.2008.17}.

\bibitem[Liu et~al.(2021)Liu, Yuan, Fu, Jiang, Hayashi, and
  Neubig]{liu2021pretrain}
Liu, P., Yuan, W., Fu, J., Jiang, Z., Hayashi, H., and Neubig, G.
\newblock Pre-train, prompt, and predict: A systematic survey of prompting
  methods in natural language processing, 2021.

\bibitem[Liu et~al.(2019)Liu, Ott, Goyal, Du, Joshi, Chen, Levy, Lewis,
  Zettlemoyer, and Stoyanov]{liu2019roberta}
Liu, Y., Ott, M., Goyal, N., Du, J., Joshi, M., Chen, D., Levy, O., Lewis, M.,
  Zettlemoyer, L., and Stoyanov, V.
\newblock Roberta: A robustly optimized bert pretraining approach.
\newblock \emph{arXiv preprint arXiv:1907.11692}, 2019.

\bibitem[Meng et~al.(2019)Meng, Liu, Zhu, Zhang, Pei, Liu, Chen, Zhang, Tao,
  Sun, and Zhou]{ijcai2019p658}
Meng, W., Liu, Y., Zhu, Y., Zhang, S., Pei, D., Liu, Y., Chen, Y., Zhang, R.,
  Tao, S., Sun, P., and Zhou, R.
\newblock Loganomaly: Unsupervised detection of sequential and quantitative
  anomalies in unstructured logs.
\newblock In \emph{Proceedings of the Twenty-Eighth International Joint
  Conference on Artificial Intelligence, {IJCAI-19}}, pp.\  4739--4745.
  International Joint Conferences on Artificial Intelligence Organization, 7
  2019.
\newblock \doi{10.24963/ijcai.2019/658}.
\newblock URL \url{https://doi.org/10.24963/ijcai.2019/658}.

\bibitem[Nedelkoski et~al.(2020)Nedelkoski, Bogatinovski, Acker, Cardoso, and
  Kao]{nedelkoski2020self}
Nedelkoski, S., Bogatinovski, J., Acker, A., Cardoso, J., and Kao, O.
\newblock Self-attentive classification-based anomaly detection in unstructured
  logs.
\newblock In \emph{2020 IEEE International Conference on Data Mining (ICDM)},
  pp.\  1196--1201. IEEE, 2020.

\bibitem[Oliner \& Stearley(2007)Oliner and Stearley]{oliner2007supercomputers}
Oliner, A. and Stearley, J.
\newblock What supercomputers say: A study of five system logs.
\newblock In \emph{37th Annual IEEE/IFIP International Conference on Dependable
  Systems and Networks (DSN'07)}, pp.\  575--584. IEEE, 2007.

\bibitem[Oprea et~al.(2015)Oprea, Li, Yen, Chin, and
  Alrwais]{oprea2015detection}
Oprea, A., Li, Z., Yen, T.-F., Chin, S.~H., and Alrwais, S.
\newblock Detection of early-stage enterprise infection by mining large-scale
  log data.
\newblock In \emph{2015 45th Annual IEEE/IFIP International Conference on
  Dependable Systems and Networks}, pp.\  45--56. IEEE, 2015.

\bibitem[Petroni et~al.(2019)Petroni, Rockt{\"a}schel, Riedel, Lewis, Bakhtin,
  Wu, and Miller]{petroni-etal-2019-language}
Petroni, F., Rockt{\"a}schel, T., Riedel, S., Lewis, P., Bakhtin, A., Wu, Y.,
  and Miller, A.
\newblock Language models as knowledge bases?
\newblock In \emph{Proceedings of the 2019 Conference on Empirical Methods in
  Natural Language Processing and the 9th International Joint Conference on
  Natural Language Processing (EMNLP-IJCNLP)}, pp.\  2463--2473, Hong Kong,
  China, November 2019. Association for Computational Linguistics.
\newblock \doi{10.18653/v1/D19-1250}.
\newblock URL \url{https://aclanthology.org/D19-1250}.

\bibitem[Prewett(2003)]{prewett2003analyzing}
Prewett, J.~E.
\newblock Analyzing cluster log files using logsurfer.
\newblock In \emph{Proceedings of the 4th Annual Conference on Linux Clusters}.
  Citeseer, 2003.

\bibitem[Radford et~al.(2019)Radford, Wu, Child, Luan, Amodei, and
  Sutskever]{radford2019language}
Radford, A., Wu, J., Child, R., Luan, D., Amodei, D., and Sutskever, I.
\newblock Language models are unsupervised multitask learners.
\newblock 2019.

\bibitem[Raffel et~al.(2020)Raffel, Shazeer, Roberts, Lee, Narang, Matena,
  Zhou, Li, and Liu]{JMLR:v21:20-074}
Raffel, C., Shazeer, N., Roberts, A., Lee, K., Narang, S., Matena, M., Zhou,
  Y., Li, W., and Liu, P.~J.
\newblock Exploring the limits of transfer learning with a unified text-to-text
  transformer.
\newblock \emph{Journal of Machine Learning Research}, 21\penalty0
  (140):\penalty0 1--67, 2020.
\newblock URL \url{http://jmlr.org/papers/v21/20-074.html}.

\bibitem[Rao et~al.(2020)Rao, Meier, Sercu, Ovchinnikov, and
  Rives]{rao2020transformer}
Rao, R., Meier, J., Sercu, T., Ovchinnikov, S., and Rives, A.
\newblock Transformer protein language models are unsupervised structure
  learners.
\newblock In \emph{International Conference on Learning Representations}, 2020.

\bibitem[Ruff et~al.(2018)Ruff, Vandermeulen, Goernitz, Deecke, Siddiqui,
  Binder, M{\"u}ller, and Kloft]{pmlr-v80-ruff18a}
Ruff, L., Vandermeulen, R., Goernitz, N., Deecke, L., Siddiqui, S.~A., Binder,
  A., M{\"u}ller, E., and Kloft, M.
\newblock Deep one-class classification.
\newblock In Dy, J. and Krause, A. (eds.), \emph{Proceedings of the 35th
  International Conference on Machine Learning}, volume~80 of \emph{Proceedings
  of Machine Learning Research}, pp.\  4393--4402. PMLR, 10--15 Jul 2018.
\newblock URL \url{https://proceedings.mlr.press/v80/ruff18a.html}.

\bibitem[Schick \& Sch{\"u}tze(2021)Schick and
  Sch{\"u}tze]{schick-schutze-2021-exploiting}
Schick, T. and Sch{\"u}tze, H.
\newblock Exploiting cloze-questions for few-shot text classification and
  natural language inference.
\newblock In \emph{Proceedings of the 16th Conference of the European Chapter
  of the Association for Computational Linguistics: Main Volume}, pp.\
  255--269, Online, April 2021. Association for Computational Linguistics.
\newblock \doi{10.18653/v1/2021.eacl-main.20}.
\newblock URL \url{https://aclanthology.org/2021.eacl-main.20}.

\bibitem[Sch\"{o}lkopf et~al.(2001)Sch\"{o}lkopf, Platt, Shawe-Taylor, Smola,
  and Williamson]{10.1162/089976601750264965}
Sch\"{o}lkopf, B., Platt, J.~C., Shawe-Taylor, J.~C., Smola, A.~J., and
  Williamson, R.~C.
\newblock Estimating the support of a high-dimensional distribution.
\newblock \emph{Neural Comput.}, 13\penalty0 (7):\penalty0 1443–1471, jul
  2001.
\newblock ISSN 0899-7667.
\newblock \doi{10.1162/089976601750264965}.
\newblock URL \url{https://doi.org/10.1162/089976601750264965}.

\bibitem[Simache \& Kaaniche(2005)Simache and Kaaniche]{1607498}
Simache, C. and Kaaniche, M.
\newblock Availability assessment of sunos/solaris unix systems based on
  syslogd and wtmpx log files: A case study.
\newblock In \emph{11th Pacific Rim International Symposium on Dependable
  Computing (PRDC'05)}, pp.\  8 pp.--, 2005.
\newblock \doi{10.1109/PRDC.2005.20}.

\bibitem[Taylor(1953)]{taylor1953cloze}
Taylor, W.~L.
\newblock “cloze procedure”: A new tool for measuring readability.
\newblock \emph{Journalism quarterly}, 30\penalty0 (4):\penalty0 415--433,
  1953.

\bibitem[Tenney et~al.(2019)Tenney, Das, and Pavlick]{tenney-etal-2019-bert}
Tenney, I., Das, D., and Pavlick, E.
\newblock {BERT} rediscovers the classical {NLP} pipeline.
\newblock In \emph{Proceedings of the 57th Annual Meeting of the Association
  for Computational Linguistics}, pp.\  4593--4601, Florence, Italy, July 2019.
  Association for Computational Linguistics.
\newblock \doi{10.18653/v1/P19-1452}.
\newblock URL \url{https://aclanthology.org/P19-1452}.

\bibitem[Vaswani et~al.(2017)Vaswani, Shazeer, Parmar, Uszkoreit, Jones, Gomez,
  Kaiser, and Polosukhin]{NIPS2017_3f5ee243}
Vaswani, A., Shazeer, N., Parmar, N., Uszkoreit, J., Jones, L., Gomez, A.~N.,
  Kaiser, L.~u., and Polosukhin, I.
\newblock Attention is all you need.
\newblock In Guyon, I., Luxburg, U.~V., Bengio, S., Wallach, H., Fergus, R.,
  Vishwanathan, S., and Garnett, R. (eds.), \emph{Advances in Neural
  Information Processing Systems}, volume~30. Curran Associates, Inc., 2017.

\bibitem[Wang et~al.(2004)Wang, Wong, and Miner]{1437839}
Wang, Y., Wong, J., and Miner, A.
\newblock Anomaly intrusion detection using one class svm.
\newblock In \emph{Proceedings from the Fifth Annual IEEE SMC Information
  Assurance Workshop, 2004.}, pp.\  358--364, 2004.
\newblock \doi{10.1109/IAW.2004.1437839}.

\bibitem[Wu et~al.(2016)Wu, Schuster, Chen, Le, Norouzi, Macherey, Krikun, Cao,
  Gao, Macherey, et~al.]{wu2016google}
Wu, Y., Schuster, M., Chen, Z., Le, Q.~V., Norouzi, M., Macherey, W., Krikun,
  M., Cao, Y., Gao, Q., Macherey, K., et~al.
\newblock Google's neural machine translation system: Bridging the gap between
  human and machine translation.
\newblock \emph{arXiv preprint arXiv:1609.08144}, 2016.

\bibitem[Xu et~al.(2009)Xu, Huang, Fox, Patterson, and Jordan]{xu2009detecting}
Xu, W., Huang, L., Fox, A., Patterson, D., and Jordan, M.~I.
\newblock Detecting large-scale system problems by mining console logs.
\newblock In \emph{Proceedings of the ACM SIGOPS 22nd symposium on Operating
  systems principles}, pp.\  117--132, 2009.

\bibitem[Yang et~al.(2019)Yang, Dai, Yang, Carbonell, Salakhutdinov, and
  Le]{yang2019xlnet}
Yang, Z., Dai, Z., Yang, Y., Carbonell, J., Salakhutdinov, R.~R., and Le, Q.~V.
\newblock Xlnet: Generalized autoregressive pretraining for language
  understanding.
\newblock \emph{Advances in neural information processing systems}, 32, 2019.

\bibitem[Yen et~al.(2013)Yen, Oprea, Onarlioglu, Leetham, Robertson, Juels, and
  Kirda]{yen2013beehive}
Yen, T.-F., Oprea, A., Onarlioglu, K., Leetham, T., Robertson, W., Juels, A.,
  and Kirda, E.
\newblock Beehive: Large-scale log analysis for detecting suspicious activity
  in enterprise networks.
\newblock In \emph{Proceedings of the 29th Annual Computer Security
  Applications Conference}, pp.\  199--208, 2013.

\bibitem[You et~al.(2022)You, Cui, Shen, Yang, Lu, Zheng, and Le]{you2022a}
You, Z., Cui, L., Shen, Y., Yang, K., Lu, X., Zheng, Y., and Le, X.
\newblock A unified model for multi-class anomaly detection.
\newblock In Oh, A.~H., Agarwal, A., Belgrave, D., and Cho, K. (eds.),
  \emph{Advances in Neural Information Processing Systems}, 2022.
\newblock URL \url{https://openreview.net/forum?id=bMYU8\_qD8PW}.

\bibitem[Zhang et~al.(2019)Zhang, Xu, Lin, Qiao, Zhang, Dang, Xie, Yang, Cheng,
  Li, et~al.]{zhang2019robust}
Zhang, X., Xu, Y., Lin, Q., Qiao, B., Zhang, H., Dang, Y., Xie, C., Yang, X.,
  Cheng, Q., Li, Z., et~al.
\newblock Robust log-based anomaly detection on unstable log data.
\newblock In \emph{Proceedings of the 2019 27th ACM Joint Meeting on European
  Software Engineering Conference and Symposium on the Foundations of Software
  Engineering}, pp.\  807--817, 2019.

\end{thebibliography}
\bibliographystyle{icml2021}
\clearpage

\end{document}